%% file: main.tex
\documentclass[10pt,twocolumn,letterpaper]{article}

\usepackage{cvpr}              %

\input{preamble}
\definecolor{cvprblue}{rgb}{0.21,0.49,0.74}
\usepackage[pagebackref,breaklinks,colorlinks,allcolors=cvprblue]{hyperref}
\usepackage{booktabs}
\usepackage{multirow}
\usepackage{multicol}
\usepackage{color, colortbl}
\usepackage{pifont}%
\usepackage{array}
\usepackage[misc]{ifsym} %
\newcommand{\cmark}{\ding{51}}%
\newcommand{\xmark}{\ding{55}}%
\newcommand{\name}{DiffusionDrive}
\newcommand{\boldparagraph}[1]{\vspace{0.1cm}\noindent{\bf #1}}

\title{\name{}: Truncated Diffusion Model for End-to-End Autonomous Driving}

\author{Bencheng Liao$^{1, 2, \diamond}$ \quad
Shaoyu Chen$^{2,3}$ \quad
Haoran Yin$^{3}$ \quad
Bo Jiang$^{2,\diamond}$ \quad
Cheng Wang$^{1,2,\diamond}$ \quad
Sixu Yan$^{2}$ \quad \\
Xinbang Zhang$^{3}$ \quad 
Xiangyu Li$^{3}$ \quad
Ying Zhang$^{3}$ \quad
Qian Zhang$^{3}$ \quad 
Xinggang Wang$^{2~\textrm{\Letter}}$
\vspace{0.3em} \\
\quad\quad\quad \textsuperscript{1} Institute of Artificial Intelligence, Huazhong University of Science \& Technology \\
\quad\quad\quad \textsuperscript{2} School of EIC, Huazhong University of Science \& Technology \\
\quad\quad\quad \textsuperscript{3} Horizon Robotics \\
Code \& Model \& Demo: \href{https://github.com/hustvl/DiffusionDrive}{ \ttfamily hustvl/DiffusionDrive}
}
\begin{document}
\maketitle

\let\thefootnote\relax\footnotetext{$^\diamond$ Intern of Horizon Robotics; $^{~\textrm{\Letter}}$ Corresponding author: Xinggang Wang (\url{xgwang@hust.edu.cn}).}

\begin{abstract}
    Recently, the diffusion model has emerged as a powerful generative technique for robotic policy learning, capable of modeling multi-mode action distributions. Leveraging its capability for end-to-end autonomous driving is a promising direction. However, the numerous denoising steps in the robotic diffusion policy and the more dynamic, open-world nature of traffic scenes pose substantial challenges for generating diverse driving actions at a real-time speed. To address these challenges, we propose a novel truncated diffusion policy that incorporates prior multi-mode anchors and truncates the diffusion schedule, enabling the model to learn denoising from anchored Gaussian distribution to the multi-mode driving action distribution. Additionally, we design an efficient cascade diffusion decoder for enhanced interaction with conditional scene context. The proposed model, DiffusionDrive, demonstrates 10$\times$ reduction in denoising steps compared to vanilla diffusion policy, delivering superior diversity and quality in just 2 steps. On the planning-oriented NAVSIM dataset, with aligned ResNet-34 backbone, DiffusionDrive achieves 88.1 PDMS without bells and whistles, setting a new record, while running at a real-time speed of 45 FPS on an NVIDIA 4090. Qualitative results on challenging scenarios further confirm that DiffusionDrive can robustly generate diverse plausible driving actions.
\end{abstract}
\input{sec/1_intro}

\input{sec/2_related}

\input{sec/3_method}

\input{sec/4_exp}

\input{sec/5_conclusion}
\section*{Acknowledgement}
We would like to acknowledge Tianheng Cheng for helpful feedback on the draft.

{
    \small
    \bibliographystyle{ieeenat_fullname}
    \bibliography{main}
}

\input{sec/X_suppl}

\end{document}

%% file: sec/1_intro.tex
\section{Introduction}
\label{sec:intro}

End-to-end autonomous driving has gained significant attention in recent years due to advancements in perception models (detection~\cite{li2022bevformer,wang2022detr3d,huang2021bevdet,polardetr}, tracking~\cite{zeng2022motr,zhang2022bytetrack,zhang2021fairmot}, online mapping~\cite{liu2023vectormapnet,liao2023maptr,liao2024maptrv2}, \etc), which directly learns the driving policy from the raw sensor inputs.
This data-driven approach offers a scalable and robust alternative to traditional rule-based motion planning, which often struggles to generalize to complex real-world driving settings.
\begin{figure}[t]
    \centering
    \includegraphics[width=\linewidth]{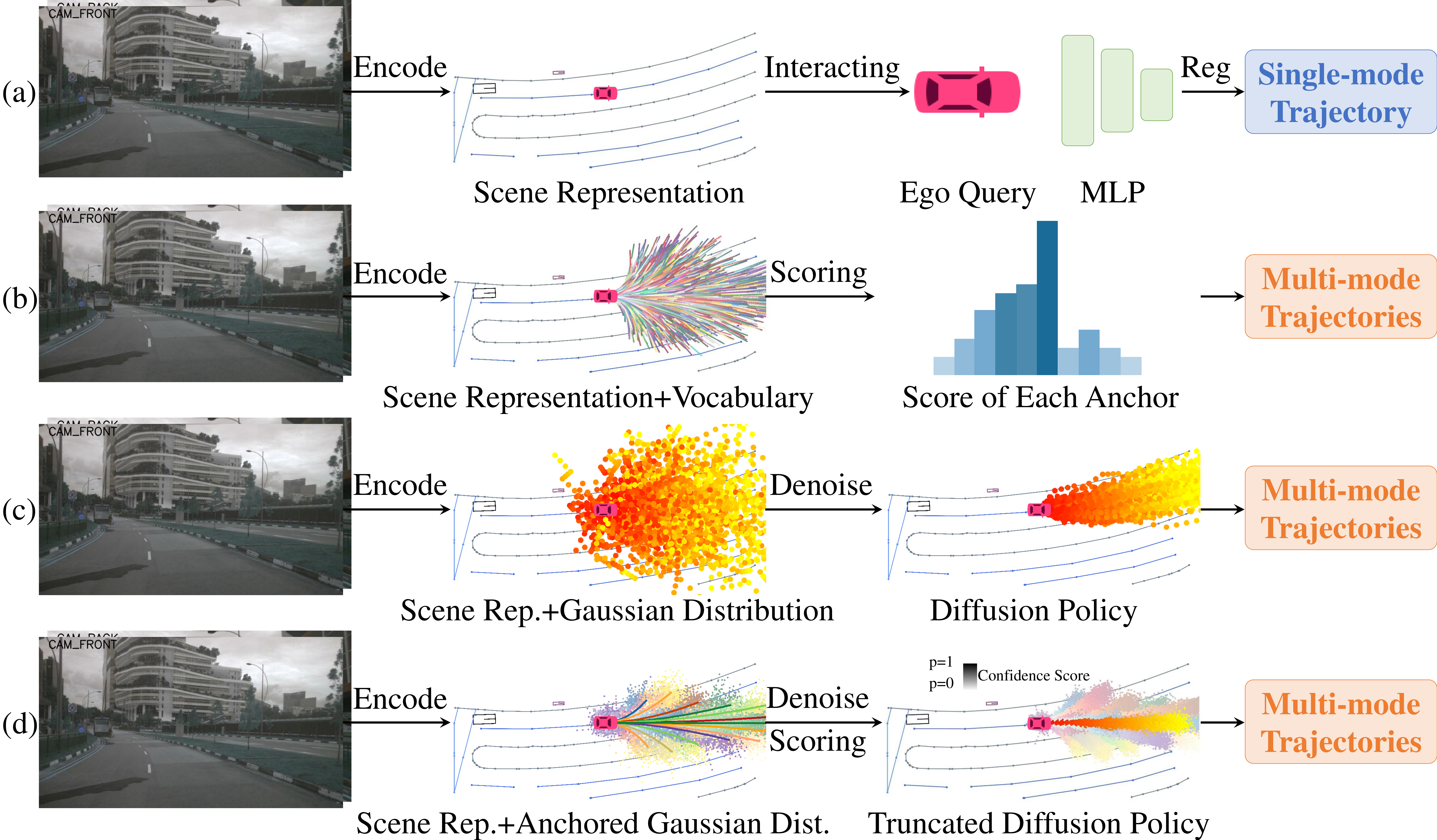}
    \vspace{-0.7cm}
    \caption{\textbf{The comparison of different end-to-end paradigms.} (a) Single mode regression~\cite{jiang2023vad,hu2023planning,transfuser}. (b) Sampling from vocabulary~\cite{vadv2,hydramdp}. (c) Vanilla diffusion policy~\cite{diffusionpolicy,janner2022diffuser}. (d) The proposed truncated diffusion policy.}
    \label{fig:intro:diffusion_decoder}
    \vspace{-0.7cm}
\end{figure}

To effectively learn from data, mainstream end-to-end planners (\eg, Transfuser~\cite{transfuser}, UniAD~\cite{hu2023planning}, VAD~\cite{jiang2023vad}) typically regress a single-mode trajectory from an ego-query as shown in Fig.~\ref{fig:intro:diffusion_decoder}a.
However, this paradigm does not account for the inherent uncertainty and multi-mode\footnote{To distinguish the term ``multimodal'' used to describe input data, we use ``multi-mode'' in this paper to refer to diverse planning decisions.} nature of driving behaviors.
Recently, VADv2~\cite{jiang2023vad} introduces a large fixed vocabulary of anchor trajectories (4096 anchors) to discretize the continuous action space and capture a broader range of driving behaviors, and then samples from these anchors based on predicted scores as shown in Fig.~\ref{fig:intro:diffusion_decoder}b.
However, this large fixed-vocabulary paradigm is fundamentally constrained by the number and quality of anchor trajectories, often failing in out-of-vocabulary scenarios.
Furthermore, managing a large number of anchors presents significant computational challenges for real-time applications.
Rather than discretizing the action space, diffusion model~\cite{diffusionpolicy} has proven to be a powerful generative decision-making policy in the robotics domain, which can directly sample multi-mode physically plausible actions from a Gaussian distribution via iterative denoising process.

\begin{figure*}[ht!]
    \centering
    \begin{subfigure}[b]{0.80\linewidth}
        \includegraphics[width=\linewidth]{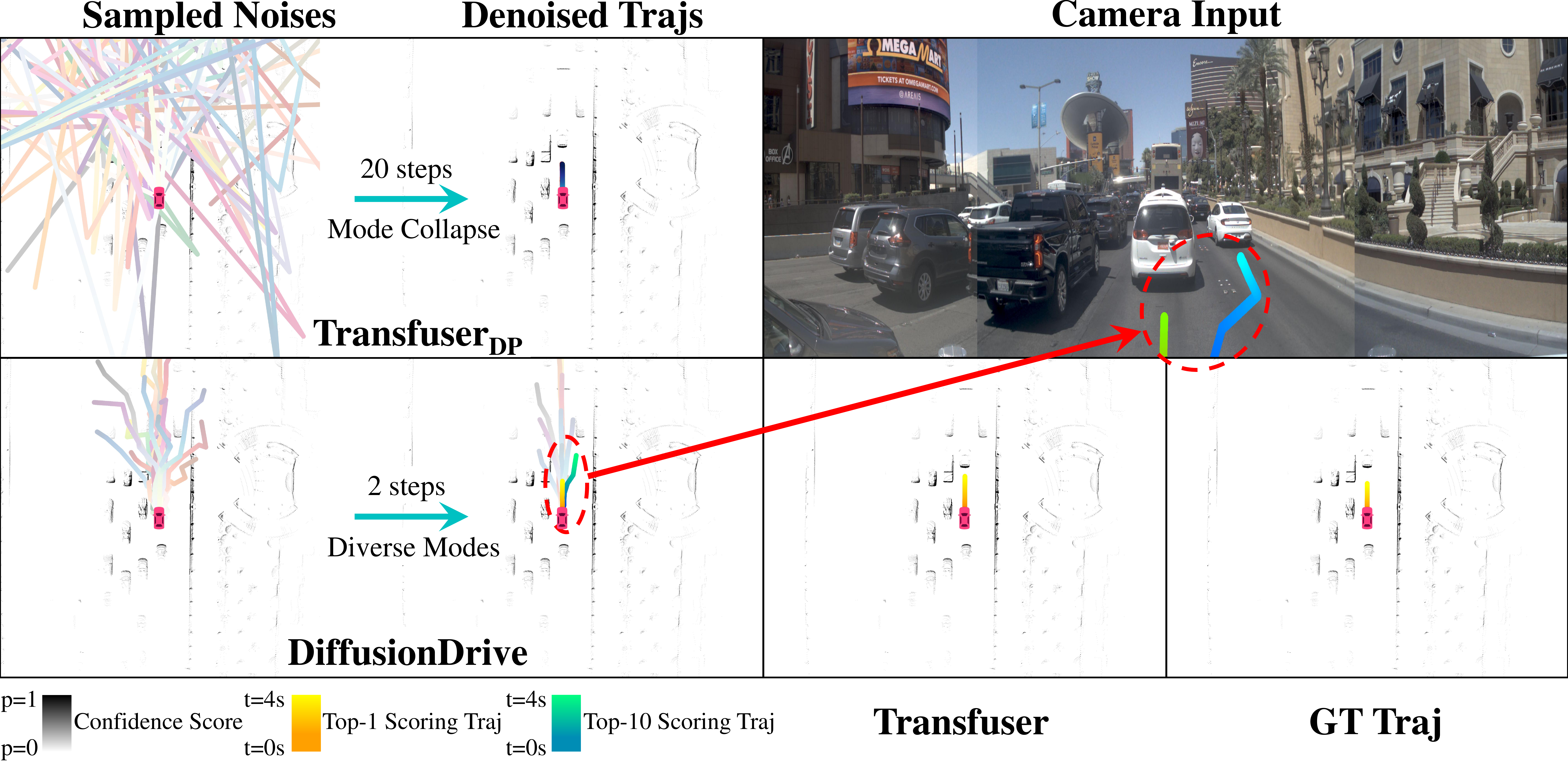}
        \caption{\textbf{Top-1's going straight and diverse top-10's lane changing.}}
        \label{fig:navsim_comp_0}
    \end{subfigure}
    \begin{subfigure}[b]{0.80\linewidth}
        \includegraphics[width=\linewidth]{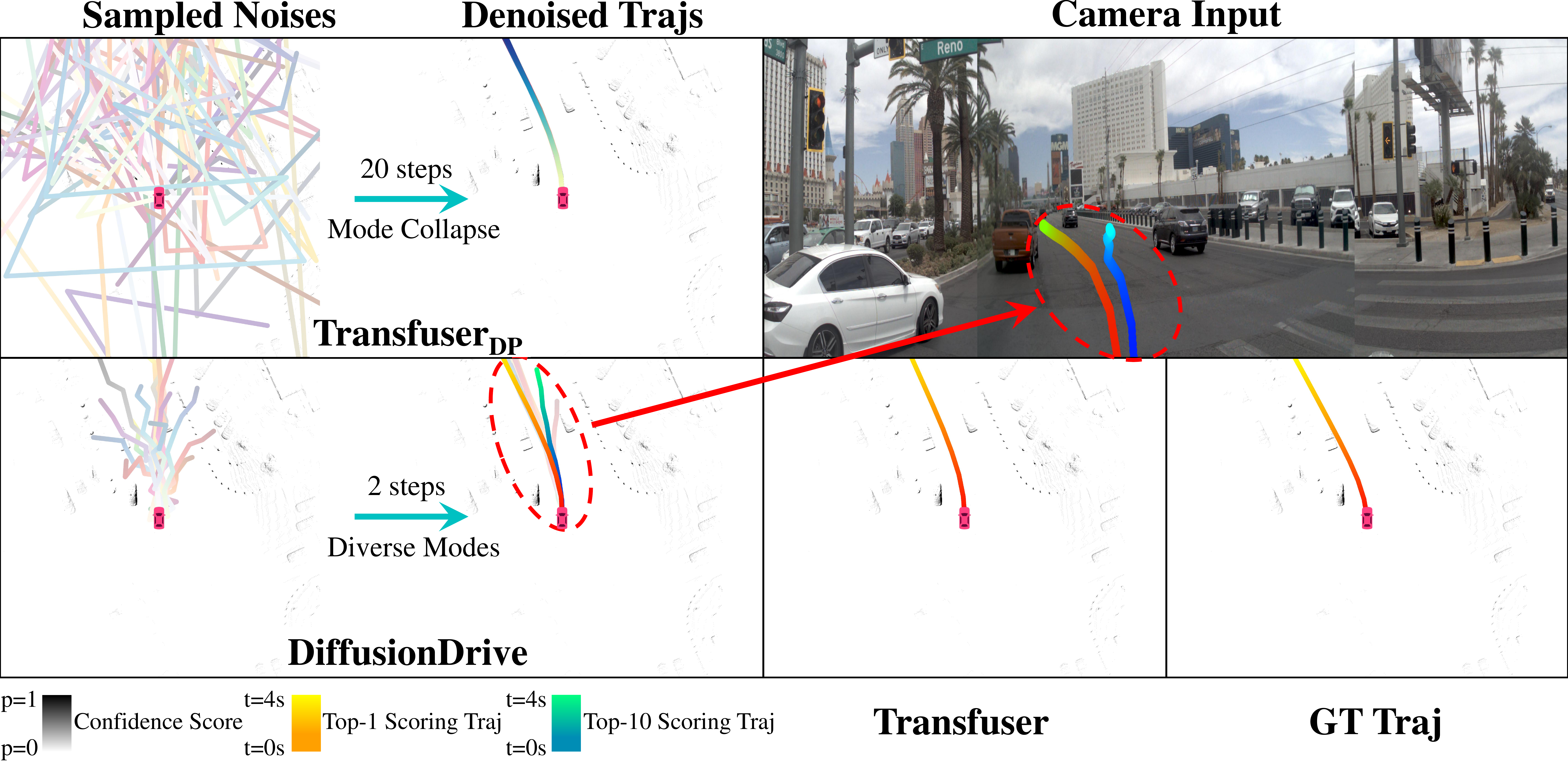}
        \caption{\textbf{Top-1's turning left and diverse top-10's lane changing.}}
        \label{fig:navsim_comp_1}
        \vspace{-0.4cm}
    \end{subfigure}
    \caption{\textbf{Qualitative comparison of Transfuser, Transfuser$_{\text{DP}}$ and \name{} on challenging scenes of NAVSIM \texttt{navtest} split.} With the same inputs from front cameras and LiDAR, \name{} achieves the highest planning quality of top-1 scoring trajectory as illustrated in Tab.~\ref{tab:roadmap}. We render the highlighted diverse trajectories predicted by \name{} in the front view. (a) and (b) shows that the top-1 scoring trajectory of \name{} closely matches the ground truth for both going straight and turning left. Additionally, \name{}’s top-10 scoring trajectory demonstrates high-quality lane changing—an ability not observed in multi-mode Transfuser$_{\text{DP}}$ and impossible for Transfuser.}
    \label{fig:navsim_comp}
    \vspace{-0.6cm}
\end{figure*}

This inspires us to replicate the success of the diffusion model in the robotics domain to end-to-end autonomous driving.
We apply the vanilla robotic diffusion policy to the well-known single-mode-regression method, Transfuser~\cite{transfuser}, by proposing a variant, Transfuser$_{\text{DP}}$, which replaces the deterministic MLP regression head with a conditional diffusion model~\cite{ronneberger2015u}.
Though Transfuser$_{\text{DP}}$ improves planning performance, two major issues arise:
\textit{1) The numerous 20 denoising steps in the vanilla DDIM diffusion policy introduce heavy computational consumption during inference as shown in Tab.~\ref{tab:roadmap}, hindering the real-time application for autonomous driving.}
\textit{2) The trajectories sampled from different Gaussian noises severely overlap with each other, as illustrated  in Fig.~\ref{fig:navsim_comp}.}
This underscores the non-trivial challenge of taming the diffusion models for the dynamic and open-world traffic scenes.

Unlike the vanilla diffusion policy, which samples actions from a random Gaussian noise conditioned on scene context, human drivers adhere to established driving patterns that they dynamically adjust in response to real-time traffic conditions.
This insight motivates us to embed these prior driving patterns into the diffusion policy by partitioning the Gaussian distribution into multiple sub-Gaussian distributions centered around prior anchors, referred to as anchored Gaussian distribution.
It is implemented by truncating the diffusion schedule to introduce a small portion of Gaussian noise around the prior anchors as shown in Fig.~\ref{fig:method_trunc}. Thanks to the multi-mode distributional expressivity of the diffusion model, the proposed truncated diffusion policy effectively covers the potential action space without requiring a large set of fixed anchors, as VADv2 does.
With more reasonable initial noise samples from the anchored Gaussian distribution, we can truncate the denoising process, reducing the required steps from 20 to just 2—a substantial speedup that satisfies the real-time requirements of autonomous driving.

To enhance the interaction with conditional scene context, we propose an efficient transformer-based diffusion decoder that interacts not only with structured queries from the perception module but also with Bird's Eye View (BEV) and perspective view (PV) features through a sparse deformable attention mechanism~\cite{zhu2021deformable}.
Additionally, we introduce a cascade mechanism to iteratively refine the trajectory reconstruction within the diffusion decoder at each denoising step.

With these innovations, we present \textbf{\name{}}, a diffusion model for real-time end-to-end autonomous driving. 
We benchmark our method on the planning-oriented NAVSIM dataset~\cite{dauner2024navsim} using  non-reactive simulation and closed-loop evaluations.
Without bells and whistles, \name{} achieves 88.1 PDMS on NAVSIM \texttt{navtest} split with the aligned ResNet-34 backbone, significantly outperforming previous state-of-the-art methods.
Even compared to the NAVSIM challenge-winning solution Hydra-MDP-$\mathcal{V}_{8192}$-W-EP~\cite{hydramdp}, which follows VADv2 with 8192 anchor trajectories and further incorporates post-processing and additional supervision, \name{} still outperforms it by 1.6 PDMS through directly learning from human demonstrations and inferring without post-processing, while running at real-time speed of 45 FPS on an NVIDIA 4090.
We further validate the superiority of \name{} on popular nuScenes dataset~\cite{caesar2020nuscenes} with open-loop evaluations, \name{} runs 1.8$\times$ faster than VAD and outperforms it~\cite{jiang2023vad} by 20.8\% lower L2 error and 63.6\% lower collision rate with the same ResNet-50 backbone, demonstrating state-of-the-art planning performance.

Our contributions can be summarized as follows:
\begin{itemize}
    \item We firstly introduce the diffusion model to the field of end-to-end autonomous driving and propose a novel truncated diffusion policy to address the issues of mode collapse and heavy computational overhead found in direct adaptation of vanilla diffusion policy to the traffic scene.
    \item We design an efficient transformer-based diffusion decoder that interacts with the conditional information in a cascaded manner for better trajectory reconstruction.
    \item Without bells and whistles, \name{} significantly outperforms previous state-of-the-art methods, achieving a record-breaking 88.1 PDMS on the NAVSIM \texttt{navtest} split with the same backbone, while maintaining real-time performance at 45 FPS on an NVIDIA 4090. 
    \item We qualitatively demonstrate that \name{} can generate more diverse and plausible trajectories, exhibiting high-quality multi-mode driving actions in various challenging scenarios.
\end{itemize}

%% file: sec/2_related.tex
\section{Related Work}
\label{sec:related}
\boldparagraph{End-to-end autonomous driving.}
UniAD~\cite{hu2023planning}, as a pioneering work, demonstrates the potential of end-to-end autonomous driving by integrating multiple perception tasks to enhance planning performance. VAD~\cite{jiang2023vad} further explores the use of compact vectorized scene representations to improve efficiency. Subsequently, a series of works~\cite{li2024enhancing,paradrive,zheng2024genad,li2024ego,wang2024driving,gu2024producing,chen2025ppad,transfuser} have adopted the single-trajectory planning paradigm to enhance planning performance further. More recently, VADv2~\cite{vadv2} shifts the paradigm towards multi-mode planning by scoring and sampling from a large fixed vocabulary of anchor trajectories. Hydra-MDP~\cite{hydramdp} improves the scoring mechanism of VADv2 by introducing extra supervision from a rule-based scorer. SparseDrive~\cite{sun2024sparsedrive} explores an alternative BEV-free solution.
Unlike existing multi-mode planning approaches, we propose a novel paradigm that leverages powerful generative diffusion models for end-to-end autonomous driving.

\boldparagraph{Diffusion model for traffic simulation.}
Driving diffusion policy has been explored in the traffic simulation by leveraging only abstract perception groundtruth~\cite{jiang2023motiondiffuser,choi2024dice,huang2024versatile,wang2024optimizing}. MotionDiffuser~\cite{jiang2023motiondiffuser} and CTG~\cite{zhong2023guided} are pioneering applications of diffusion models for multi-agent motion prediction, using a conditional diffusion model to sample target trajectories from Gaussian noise. CTG++~\cite{zhong2023languageguided} further incorporates a large language model (LLM) for language-driven guidance, improving usability and enabling realistic traffic simulations. Diffusion-ES~\cite{yang2024diffusion} replaces reward-gradient-guided denoising with evolutionary search. 
Moving beyond diffusion models limited to traffic simulation with perception groundtruth, our approach unlocks the potential of diffusion models for real-time, end-to-end autonomous driving through our proposed truncated diffusion policy and efficient diffusion decoder.

\begin{figure*}[htbp!]
    \centering
    \includegraphics[width=0.82\linewidth]{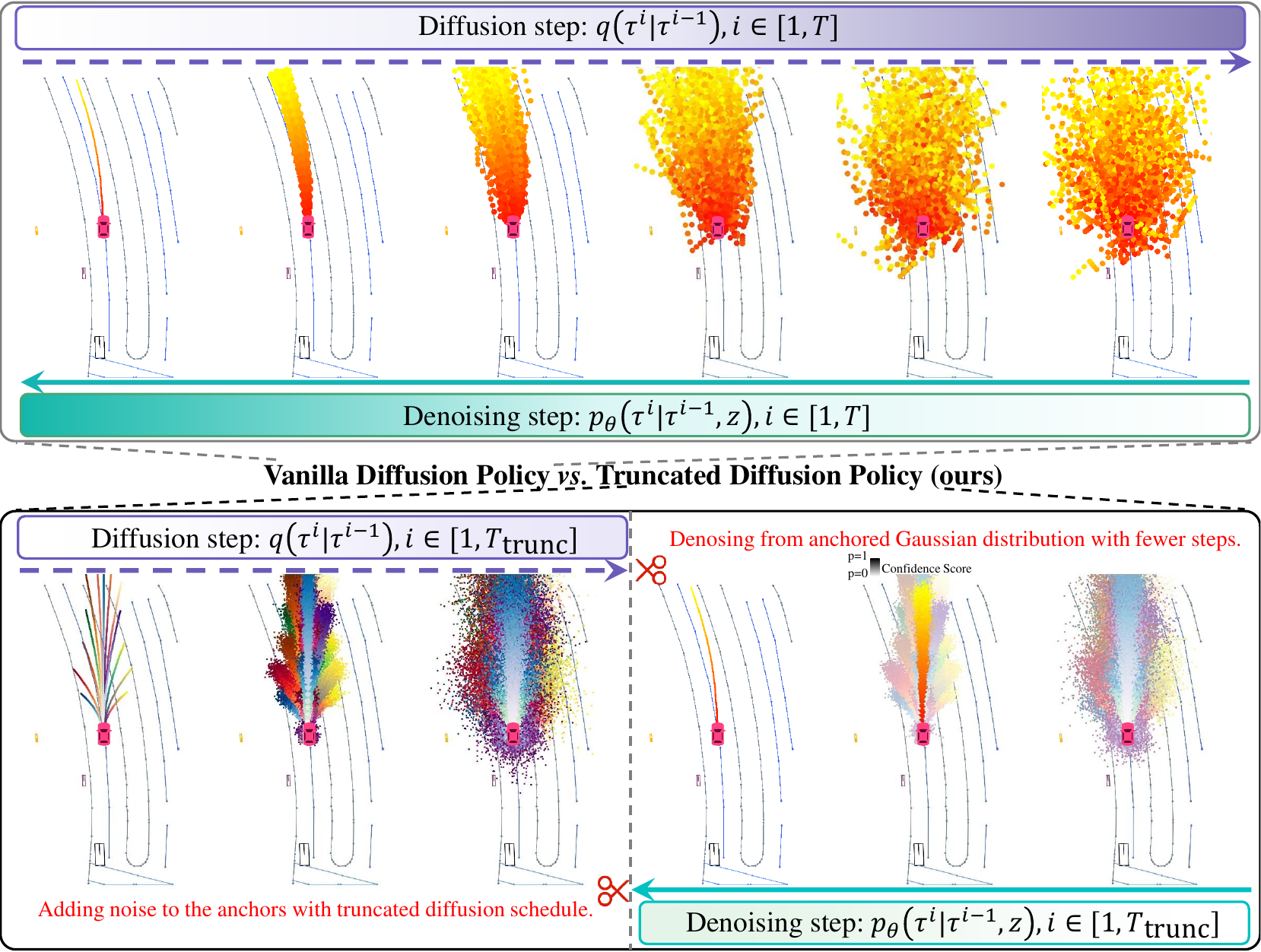}
    \vspace{-0.2cm}
    \caption{\textbf{Illustration of truncated diffusion policy by comparing with vanilla diffusion policy.} We truncate the diffusion process and only add a small portion of Gaussian noise to diffuse the anchor trajectories. Then, we train the diffusion model to reconstruct the ground-truth trajectory from the anchored Gaussian distribution with conditional scene context. During the inference, we also truncate the denoising process by starting from the better samples in the anchored Gaussian distribution than the pure Gaussian noise.}
    \label{fig:method_trunc}
    \vspace{-0.6cm}
\end{figure*}

\boldparagraph{Diffusion model for robotic policy learning.}
Diffusion policy~\cite{diffusionpolicy} demonstrates the great potential in robotic policy learning, effectively capturing multi-mode action distributions and high-dimensional action spaces. Diffuser~\cite{janner2022diffuser} proposes an unconditional diffusion model for trajectory sampling, incorporating techniques such as classifier-free guidance and image inpainting to achieve guided sampling. Subsequently, numerous works have applied diffusion models to various robotic tasks, including stationary manipulation~\cite{ze20243d,ajay2023is}, mobile manipulation~\cite{yan2024m2diffuser}, autonomous navigation~\cite{sridhar2024nomad,yu2024ldp}, quadruped locomotion~\cite{stamatopoulou2024dippest}, and dexterous manipulation~\cite{weng2024dexdiffuser}.
However, directly applying vanilla diffusion policy to end-to-end autonomous driving poses unique challenges, as it requires real-time efficiency and the generation of plausible multi-mode trajectories in dynamic and open-world traffic scenes. In this work, we propose a novel truncated diffusion policy to address these challenges, introducing concepts that have not yet been explored in the robotics field.

\boldparagraph{Diffusion model for image generation.}
Diffusion models have been extensively adopted for image generation tasks~\cite{vavae,fasterdit,rombach2022high,song2021scorebased,zhu2024dig}. DDIM~\cite{song2020denoising} enhances DDPM~\cite{ho2020denoising} by enabling efficient sampling with significantly fewer steps based on non-Markovian diffusion processes. Flow matching~\cite{lipman2023flow,liu2022flow} further optimizes the generative process by directly modeling continuous probability flows. TDPM~\cite{zheng2023truncated} proposes truncated denoising, which initiates the generation process from an implicit intermediate distribution to accelerate sampling. 
In contrast to these approaches, our method introduces an explicit driving prior within the diffusion policy, effectively guiding the diffusion process toward more accurate and efficient generation tailored specifically for end-to-end autonomous driving.

%% file: sec/3_method.tex
\section{Method}

\begin{figure*}[t!]
    \centering
    \includegraphics[width=1.0\linewidth]{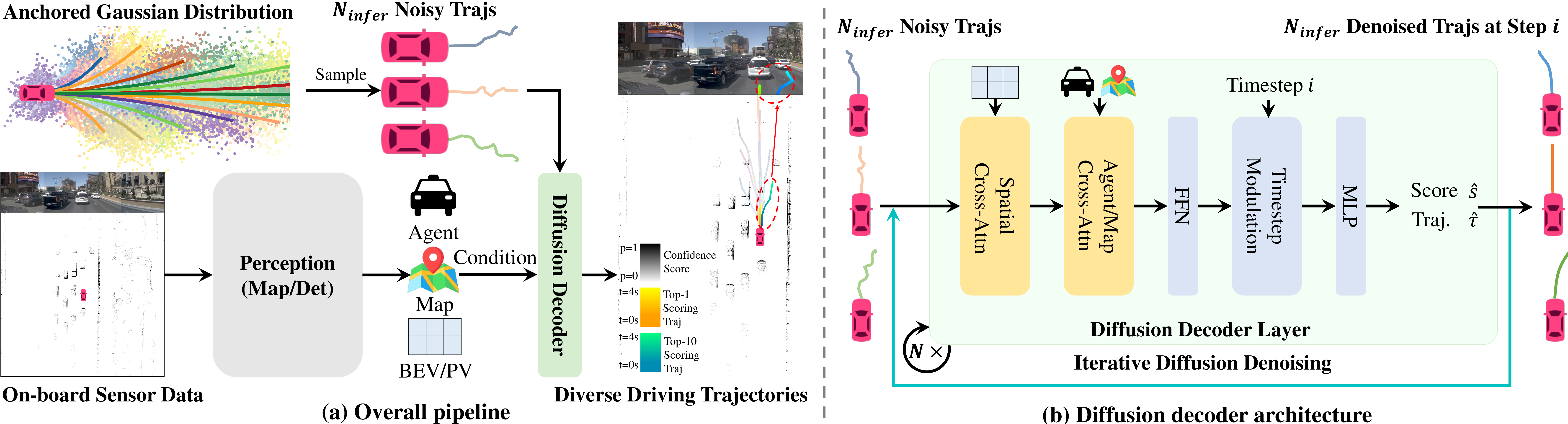}
    \vspace{-0.6cm}
    \caption{\textbf{Overall architecture of \name{}.} (a) \name{} can integrate various existing perception modules and sensor inputs. (b) The designed diffusion decoder takes the sampled noisy trajectories from anchored Gaussian distribution as input and progressively denoises them with enhanced interactions with the conditional scene context in a cascade manner to generate the final predictions.}
    \label{fig:method_pipeline}
    \vspace{-0.6cm}
\end{figure*}
\subsection{Preliminary}
\boldparagraph{Task formulation.}
End-to-end autonomous driving takes raw sensor data as input and predicts the future trajectory of the ego-vehicle.
The trajectory is represented as a sequence of waypoints $\tau = \{(x_t, y_t)\}_{t=1}^{T_f}$, where $T_f$ denotes the planning horizon, and $(x_t, y_t)$ is the location of each waypoint at time $t$ in the current ego-vehicle coordinate system.

\boldparagraph{Conditional diffusion model.}
The conditional diffusion model poses a forward diffusion process as gradually adding noise to the data sample, which can be defined as:
\begin{equation}
    q\left(\tau^i\mid \tau^0\right)=\mathcal{N}\left(\tau^i;\sqrt{\bar{\alpha}^i}\boldsymbol{\tau}^0,\left(1-\bar{\alpha}^i\right)\mathbf{I}\right),
\end{equation}
where $\tau^0$ is the clean data sample, and $\tau^i$ is the data sample with noise at time $i$ (Note: we use superscript $i$ to denote diffusion timestep).
The constant $\bar{\alpha^i}=\prod_{s=1}^i\alpha^s=\prod_{s=1}^i(1-\beta^s)$ and $\beta^s$ is the noise schedule. We train the reverse process model $f_\theta(\tau^i,z,i)$ to predict $\tau^0$ from $\tau^i$ with the guidance of conditional information $z$, where $\theta$ is the trainable model parameter.
During inference, the trained diffusion model $f_\theta$ progressively refines from the random noise $\tau^T$ sampled in Gaussian distribution to the predicted clean data sample $\tau^0$ with the guidance of conditional information $z$, which is defined as:
\begin{equation}
    p_\theta\left(\boldsymbol{\tau}^{0} \mid z\right)=\int p\left(\boldsymbol{\tau}^T\right) \prod_{i=1}^T p_\theta\left(\boldsymbol{\tau}^{i-1} \mid \boldsymbol{\tau}^i, z\right)\mathrm{d} \boldsymbol{\tau}^{1:T} .
\end{equation}

\subsection{Investigation}
\label{sec:investigation}
\boldparagraph{Turn Transfuser~\cite{transfuser} into conditional diffusion model.}
We begin from the representative deterministic end-to-end planner Transfuser \cite{transfuser} and turn it into a generative model Transfuser$_{\text{DP}}$ by simply replacing the regression MLP layers with the conditional diffusion model UNet following vanilla diffusion policy \cite{diffusionpolicy}.
During the evaluation, we sample a random noise and progressively refine it with 20 steps.
Tab.~\ref{tab:roadmap} shows that Transfuser$_{\text{DP}}$ achieves better planning quality than deterministic Transfuser.

\boldparagraph{Mode collapse.}
To further investigate the multi-mode property of the vanilla diffusion policy in driving, we sampled 20 random noises from Gaussian distribution and denoised them using 20 steps. As shown in Fig.~\ref{fig:navsim_comp}, the different random noises converge to similar trajectories after the denoising process.
To quantitatively analyze the phenomenon of mode collapse, we define a mode diversity score $\mathcal{D}$ based on the mean Intersection over Union (mIoU) between each denoised trajectory and the union of all denoised trajectories:
\begin{equation}
    \mathcal{D} = 1 - \frac{1}{N}\sum_{i=1}^N \frac{\text{Area}(\tau_i \cap \bigcup_{j=1}^N \tau_j)}{\text{Area}(\tau_i \cup \bigcup_{j=1}^N \tau_j)},
    \label{eq:div}
\end{equation}
where $\tau_i$ represents the $i$-th denoised trajectory, $N$ is the total number of sampled trajectories and $\bigcup_{j=1}^N \tau_j$ is the union of all denoised trajectories. A higher mIoU indicates less diversity of the denoised trajectories.
The quantitative mode diversity results in Tab.~\ref{tab:roadmap} further validate the observations presented in Fig.~\ref{fig:navsim_comp}.

\boldparagraph{Heavy denoising overhead.}
The DDIM~\cite{song2020denoising} diffusion policy requires 20 denoising steps to transform random noise into a feasible trajectory, which introduces significant computational overhead, reducing the FPS from 60 to 7, as shown in Tab.~\ref{tab:roadmap}, and making it impractical for real-time online driving applications.
\begin{table*}[htbp!]
    \centering
    \renewcommand\tabcolsep{4.3pt}
    \input{tables/tab1}
    \vspace{-0.3cm}
    \caption{\textbf{Comparison on planning-oriented NAVSIM \texttt{navtest} split with closed-loop metrics.} ``C \& L'' denotes the use of both camera and LiDAR as sensor inputs. ``$\mathcal{V}_{8192}$'' denotes 8192 anchors. ``Hydra-MDP-$\mathcal{V}_{8192}$-W-EP'' is a variant of Hydra-MDP~\cite{hydramdp}, which is further trained to fit the EP evaluation metric with additional supervision from the rule-based evaluator and uses weighted confidence post-processing. \name{} simply learns from human demonstrations and infers without post-processing. The \textbf{best} and the \underline{second best} results are denoted by \textbf{bold} and \underline{underline}.}
    \label{tab:main_navsim}
\end{table*}

\begin{table*}[htbp!]
    \vspace{-0.4cm}
    \centering
    \renewcommand\tabcolsep{2.8pt}
    \begin{tabular}{l|cccccl|lcrr|c|rr}
        \toprule
        \multirow{2}{*}{Method}&\multirow{2}{*}{NC$\uparrow$} &\multirow{2}{*}{DAC$\uparrow$} & \multirow{2}{*}{TTC$\uparrow$} & \multirow{2}{*}{Comf.$\uparrow$} & \multirow{2}{*}{EP$\uparrow$} & \cellcolor{gray!30}& \multicolumn{4}{c|}{Plan Module Time} & \multirow{2}{*}{$\mathcal{D}\uparrow$}& \multirow{2}{*}{Para.$\downarrow$}& \multirow{2}{*}{FPS$\uparrow$} \\
        & & & & & & \cellcolor{gray!30}\multirow{-2}{*}{PDMS$\uparrow$}& Arch.& Step Time$\downarrow$& Steps $\downarrow$ & Total $\downarrow$ &  \\
        \midrule
        Transfuser &97.7 & 92.8 & 92.8 & \textbf{100} & 79.2 & \cellcolor{gray!30}84.0 &MLP  & \textbf{0.2ms} &\textbf{1}  & \textbf{0.2ms}&   0\%& \textbf{56M}& \textbf{60} \\ 
        Transfuser$_{\text{DP}}$ & 97.5&93.7&92.7&\textbf{100}&79.4&\cellcolor{gray!30}84.6$_{\textbf{\text{\color{ForestGreen}+0.6}}}$ & UNet& 6.5ms&20 &130.0ms & 11\%& 101M& 7 \\ 
        Transfuser$_{\text{TD}}$ & \underline{97.9}&\underline{94.2}&\underline{93.9}&\textbf{100}&\underline{80.2}&\cellcolor{gray!30}\underline{85.7}$_{\textbf{\text{\color{ForestGreen}+1.7}}}$&UNet& 6.9ms&2 & 13.8ms & \underline{70\%} & 102M& 27 \\ 
        \name{} & \textbf{98.2}&\textbf{96.2}&\textbf{94.7}&\textbf{100}&\textbf{82.2}&\cellcolor{gray!30}\textbf{88.1}$_{\textbf{\text{\color{ForestGreen}+4.1}}}$&Dec.& \underline{3.8ms}&\underline{2}  & \underline{7.6ms} & \textbf{74\%} & \underline{60M}& \underline{45} \\ 
        \bottomrule
    \end{tabular}
    \vspace{-0.3cm}
    \caption{\textbf{Roadmap from Transfuser to \name{} on NAVSIM \texttt{navtest} split.} ``Transfuser$_{\text{DP}}$'' denotes Transfuser with vanilla DDIM diffusion policy~\cite{diffusionpolicy}. ``Transfuser$_{\text{TD}}$'' denotes Transfuser with truncated diffusion policy. ``Step Time'' denotes the runtime of each denoising step. ``FPS'' and runtime are measured on an NVIDIA 4090 GPU. ``$\mathcal{D}$'' denotes the mode diversity score defined in Eq.~\eqref{eq:div}.}
    \label{tab:roadmap}
    \vspace{-0.6cm}
\end{table*}

\subsection{Truncated Diffusion}
\label{sec:truncated}
Human driving follows fixed patterns, unlike the random noise denoising in vanilla diffusion policy. Motivated by this, we propose a truncated diffusion policy that begins the denoising process from an anchored Gaussian distribution instead of a standard Gaussian distribution. To enable the model to learn to denoise from the anchored Gaussian distribution to the desired driving policy, we further truncate the diffusion schedule during training, adding only a small amount of Gaussian noise to the anchors.

\boldparagraph{Training.}
We first construct the diffusion process by adding Gaussian noise to anchors $\{\mathbf{a}_k\}_{k=1}^{N_\text{anchor}}$ clustered by K-Means on the training set, where $\mathbf{a}_k=\{(x_t, y_t)\}_{t=1}^{T_f}$.
We truncate the diffusion noise schedule to diffuse the anchors to the anchored Gaussian distribution:
\begin{equation}
    \tau_k^i = \sqrt{\bar{\alpha}^i}\mathbf{a}_k + \sqrt{1-\bar{\alpha}^i}\boldsymbol{\epsilon}, \quad \boldsymbol{\epsilon} \sim \mathcal{N}(0, \mathbf{I}),
\end{equation}
where $i \in [1,T_\text{trunc}]$ and $T_\text{trunc} \ll T$ is the truncated diffusion steps.

During training, the diffusion decoder $f_\theta$ takes as input $N_\text{anchor}$ noisy trajectories $\{\tau_k^i\}_{k=1}^{N_\text{anchor}}$ and predicts classification scores $\{\hat{s}_k\}_{k=1}^{N_\text{anchor}}$ and denoised trajectories $\{\hat{\tau}_k\}_{k=1}^{N_\text{anchor}}$:
\begin{equation}
    \{\hat{s}_k, \hat{\tau}_k\}_{k=1}^{N_\text{anchor}} = f_\theta(\{\tau_k^i\}_{k=1}^{N_\text{anchor}}, z),
\end{equation}
where $z$ represents the conditional information. We assign the noisy trajectory around the closest anchor to the ground truth trajectory $\tau_\text{gt}$ as positive sample ($y_k=1$) and others as negative samples ($y_k=0$). The training objective combines trajectory reconstruction and classification:
\begin{equation}
    \mathcal{L} = \sum_{k=1}^{N_\text{anchor}} [y_k \mathcal{L}_{\text{rec}}(\hat{\tau}_k, \tau_\text{gt}) + \lambda\text{BCE}(\hat{s}_k, y_k)],
\end{equation}
where $\lambda$ balances the simple L1 reconstruction loss $\mathcal{L}_{\text{rec}}$ and binary cross-entropy (BCE) classification loss.

\boldparagraph{Inference.}
We use a truncated denoising process that starts with noisy trajectories sampled from the anchored Gaussian distribution and progressively denoises them to final predictions. At each denoising timestep, the estimated trajectories from the previous step are passed to the diffusion decoder $f_\theta$, which predicts classification scores $\{\hat{s}_k\}_{k=1}^{N_\text{infer}}$ and coordinates $\{\hat{\tau}_k\}_{k=1}^{N_\text{infer}}$. After obtaining the current timestep's predictions, we apply the DDIM~\cite{song2020denoising} update rule to sample trajectories for the next timestep.

\boldparagraph{Inference flexibility.}
A key advantage of our approach lies in its inference flexibility. While the model is trained with $N_\text{anchor}$ trajectories, the inference process can accommodate an arbitrary number of trajectory samples $N_\text{infer}$, where $N_\text{infer}$ can be dynamically adjusted based on computational resources or application requirements.

\subsection{Architecture}
The overall architecture of our proposed method, \name{}, is illustrated in Fig.~\ref{fig:method_pipeline}.
\name{} can integrate various existing perception modules used in previous end-to-end planners~\cite{hu2023planning,jiang2023vad,transfuser,sun2024sparsedrive} and take different sensor inputs.
The designed diffusion decoder is tailored for the complex and challenging driving application, which has enhanced interactions with the conditional scene context.

\boldparagraph{Diffusion decoder.}
Given the set of sampled noisy trajectories $\{\hat{\tau}_k\}_{k=1}^{N_\text{infer}}$ from the anchored Gaussian distribution, we begin by applying deformable spatial cross-attention~\cite{zhu2021deformable,wang2022detr3d,lin2022sparse4d} to interact with Bird's Eye View (BEV) or Perspective View (PV) features based on the trajectory coordinates. Subsequently, cross-attention is performed between the trajectory features and the agent/map queries derived from the perception module, followed by a feed-forward network (FFN).
To encode the diffusion timestep information, we utilize a Timestep Modulation layer, which is followed by a Multi-Layer Perceptron (MLP) that predicts the confidence score and the offset relative to the initial noisy trajectory coordinates. The output from this diffusion decoder layer serves as the input for the subsequent cascade diffusion decoder layer.
\name{} further reuses the cascade diffusion decoder to iteratively denoise the trajectory during inference, with parameters shared across the different denoising timesteps. The final trajectory with the highest confidence score is selected as the output.

%% file: tables/tab1.tex
\begin{tabular}{l|ccr|cccccc}
    \toprule
    Method & Input & Img. Backbone & Anchor& NC $\uparrow$ &DAC $\uparrow$ & TTC $\uparrow$& Comf. $\uparrow$ & EP $\uparrow$ & \cellcolor{gray!30}PDMS $\uparrow$  \\
    \midrule
    UniAD~\cite{hu2023planning} & Camera & ResNet-34~\cite{he2016deep} & 0 & 97.8 & 91.9 & 92.9 & \textbf{100} & 78.8 & \cellcolor{gray!30}83.4 \\
    PARA-Drive~\cite{paradrive} & Camera & ResNet-34~\cite{he2016deep} & 0 & 97.9 & 92.4 & 93.0 & 99.8 & 79.3 & \cellcolor{gray!30}84.0 \\
    LTF~\cite{transfuser} & Camera & ResNet-34~\cite{he2016deep} & 0 & 97.4 & 92.8 & 92.4 & \textbf{100} & 79.0 & \cellcolor{gray!30}83.8 \\
    Transfuser~\cite{transfuser} & C \& L & ResNet-34~\cite{he2016deep} & 0 & 97.7 & 92.8 & 92.8 & \textbf{100} & 79.2 & \cellcolor{gray!30}84.0 \\
    DRAMA~\cite{yuan2024drama} & C \& L & ResNet-34~\cite{he2016deep} & 0 & 98.0 & 93.1 & \textbf{94.8} & \textbf{100} & \underline{80.1} & \cellcolor{gray!30}85.5 \\
    VADv2-$\mathcal{V}_{8192}$~\cite{vadv2} & C \& L & ResNet-34~\cite{he2016deep} & 8192 & 97.2 & 89.1 & 91.6 & \textbf{100} & 76.0 & \cellcolor{gray!30}80.9 \\
    Hydra-MDP-$\mathcal{V}_{8192}$~\cite{hydramdp} & C \& L & ResNet-34~\cite{he2016deep} & 8192 & 97.9 & 91.7 & 92.9 & \textbf{100} & 77.6 & \cellcolor{gray!30}83.0 \\
    Hydra-MDP-$\mathcal{V}_{8192}$-W-EP~\cite{hydramdp} & C \& L & ResNet-34~\cite{he2016deep} & 8192 & \textbf{98.3}&\underline{96.0}&94.6&\textbf{100}&78.7&\cellcolor{gray!30}86.5 \\
    
    \midrule
    \name{} (Ours) & C \& L & ResNet-34~\cite{he2016deep} & 20 & \underline{98.2}  & \textbf{96.2}  & \underline{94.7}  & \textbf{100}  & \textbf{82.2}  & \cellcolor{gray!30}\textbf{88.1}\\

    \bottomrule
\end{tabular}%

%% file: sec/4_exp.tex
\section{Experiment}

\begin{table*}[htbp!]
    \begin{center}
    \centering
    \renewcommand\tabcolsep{3.9pt}
    \begin{tabular}{c|ccccc|c|cccccc}
    \toprule
    \multirow{2}{*}{ID} & UNet & Ego Query & Spatial & Agent/Map & Cascade  & \multirow{2}{*}{Param.$\downarrow$} & \multicolumn{6}{c}{Planning Metric}\\
    & Decoder & Interaction & Cross-attn & Cross-attn & Decoder & & NC$\uparrow$ & DAC$\uparrow$ & TTC$\uparrow$ & Comf.$\uparrow$ & EP$\uparrow$ &\cellcolor{gray!30} PDMS$\uparrow$ \\
    \midrule
    1 &\cmark&\cmark & \xmark & \xmark & \xmark &102M & 97.9&94.2&93.9&100&80.2&\cellcolor{gray!30}85.7 \\
    \midrule
    2 &\xmark&\cmark&\xmark & \xmark & \xmark & 57M & 88.7 & 83.2 & 80.0 & 84.8 & 43.3 & \cellcolor{gray!30}55.1 \\
    3 &\xmark&\cmark & \cmark & \xmark & \xmark & 58M & 98.2&95.4&94.4&100&81.3&\cellcolor{gray!30}87.1\\
    4 &\xmark&\cmark & \xmark & \cmark & \xmark & 58M&97.9&93.5&93.8&100&79.8&\cellcolor{gray!30}85.1 \\
    5 &\xmark&\cmark & \cmark & \cmark & \xmark & 59M&98.0&95.8&94.4&100&81.7&\cellcolor{gray!30}87.4\\
    6 &\xmark&\cmark & \cmark & \cmark & \cmark & 60M&98.2  &96.2  &94.7  &100  &82.2  & \cellcolor{gray!30}88.1\\
    \bottomrule
    \end{tabular}
    \end{center}
    \vspace{-0.6cm}
    \caption{\textbf{Ablation for design choices.} ``Cascade Decoder'' indicates that we stack 2 cascade diffusion decoder layers. ID-1 refers to Transfuser$_{\text{TD}}$ in Tab.~\ref{tab:roadmap}, utilizing conditional UNet and interaction with the ego-query, which Transfuser uses to directly regress the single-mode trajectory.}
    \label{tab:design}
    \vspace{-0.3cm}
    \end{table*}

    \begin{table*}[ht!]
        \centering
        \begin{minipage}[t]{0.32\linewidth}
            \centering
            \renewcommand\arraystretch{1.1}
            \renewcommand\tabcolsep{3.5pt}
            \small
            \scalebox{0.73}{
            \begin{tabular}{l|c|cccccc}
                \toprule
                Steps & Param. & NC & DAC & TTC & Comf. & EP & \cellcolor{gray!30}PDMS \\
                \midrule
                1 &60M  & 98.3 & 96.0 & 94.7 & 100 & 82.1 & \cellcolor{gray!30}87.9 \\ 
                2 &60M  &98.2  &96.2  &94.7  &100  &82.2  & \cellcolor{gray!30}88.1 \\ 
                3 &60M  & 98.2  &96.3  &   94.7 & 100 & 92.2  & \cellcolor{gray!30}88.1 \\ 
    
                \bottomrule
            \end{tabular}}
            \vspace{-0.3cm}
            \caption{\textbf{Denoising step number.}}
            \label{tab:step}
        \end{minipage}
        \hfill
        \begin{minipage}[t]{0.32\linewidth}
            \centering
            \renewcommand\arraystretch{1.1}
            \renewcommand\tabcolsep{3.5pt}
            \small
            \scalebox{0.73}{
            \begin{tabular}{l|c|cccccc}
                \toprule
                Stages & Param. & NC & DAC & TTC & Comf. & EP & \cellcolor{gray!30}PDMS \\
                \midrule
                1 &59M & 98.0 & 95.8 & 94.4 & 100 & 81.7 & \cellcolor{gray!30}87.4 \\ 
                2 &60M  &98.2  &96.2  &94.7  &100  &82.2  & \cellcolor{gray!30}88.1\\ 
                4 &65M & 98.4 & 96.2 & 94.9 & 100 & 82.4 & \cellcolor{gray!30}88.2 \\ 
                \bottomrule
            \end{tabular}}
            \vspace{-0.3cm}
            \caption{\textbf{Cascade stages.}}
            \label{tab:stage}
        \end{minipage}
        \hfill
        \begin{minipage}[t]{0.32\linewidth}
            \centering
            \renewcommand\arraystretch{1.1}
            \renewcommand\tabcolsep{3.5pt}
            \small
            \scalebox{0.73}{
            \begin{tabular}{l|c|cccccc}
                \toprule
                $N_{\text{infer}}$ & Param. & NC & DAC & TTC & Comf. & EP & \cellcolor{gray!30}PDMS \\
                \midrule
                10 & 60M & 97.9 & 93.5 & 93.1 & 100 & 80.0 & \cellcolor{gray!30}84.9 \\ 
                20 & 60M  &98.2  &96.2  &94.7  &100  &82.2  & \cellcolor{gray!30}88.1 \\ 
                40 & 60M & 98.5  & 96.2 &  94.8& 100 & 82.5 & \cellcolor{gray!30}88.2 \\ 
                \bottomrule
            \end{tabular}}
            \vspace{-0.3cm}
            \caption{\textbf{Number of sampled noises $N_{\text{infer}}$.}}
            \label{tab:sampled}
        \end{minipage}
        \vspace{-0.4cm}
    \end{table*}

\subsection{Dataset}

\boldparagraph{NAVSIM.}
The NAVSIM dataset~\cite{dauner2024navsim} is a real-world planning-oriented dataset builds upon OpenScene~\cite{openscene2023}, a compact redistribution of nuPlan~\cite{nuplan}, the largest publicly available annotated driving dataset. NAVSIM leverages eight cameras to achieve a full 360$^\circ$ FOV, along with a merged LiDAR point cloud derived from five sensors. Annotations are provided at a frequency of 2Hz and include both HD maps and object bounding boxes. The dataset is designed to emphasize challenging driving scenarios involving dynamic changes in driving intentions, while deliberately excluding trivial situations such as stationary scenes or constant-speed driving.

NAVSIM benchmarks planning performance using non-reactive simulations and closed-loop metrics for comprehensive evaluation.  In this paper, we employ the proposed PDM score (PDMS)~\cite{dauner2024navsim}, which is a weighted combination of several sub-scores: no at-fault collisions (NC), drivable area compliance (DAC), time-to-collision (TTC), comfort (Comf.), and ego progress (EP).

\subsection{Implementation Detail}
We adopt the same perception modules and ResNet-34 backbone~\cite{he2016deep} as Transfuser for fair comparison.
In the diffusion decoder layer, we employ spatial cross-attention to only interact with BEV features following Transfuser's BEV-based setting. We only perform agent cross-attention, since the perception module of Transfuser does not include vectorized map construction.
We stack 2 cascade diffusion decoder layers and apply truncated diffusion policy with 20 clustered anchors. The training diffusion schedule is truncated by 50/1000 to diffuse the anchors, while during inference, we use only 2 denoising steps and select the top-1 scoring predicted trajectory for evaluation.
The training and inference recipe directly follows Transfuser: We use three cropped and downscaled forward-facing camera images, concatenated as a 1024$\times$256 image, and a rasterized BEV LiDAR as input;
\name{} is trained on \texttt{navtrain} split from scratch for 100 epochs with AdamW optimizer on 8 NVIDIA 4090 GPUs with total batch size of 512, setting the learning rate to $6\times 10^{-4}$.
No test-time augmentation is applied and the final output for evaluation on \texttt{navtest} split is 8-waypoint trajectory over 4 seconds.

\subsection{Quantitative Comparison}
Tab.~\ref{tab:main_navsim} compares \name{} with state-of-the-art methods on NAVSIM \texttt{navtest} split.
With the same ResNet-34 backbone, \name{} achieves 88.1 PDMS score, demonstrating significant superior performance over the previous learning-based methods. 
Compared to VADv2, \name{} surpasses it by 7.2 PDMS while reducing the number of anchors from 8192 to 20, representing a 400$\times$ reduction. 
\name{} also outperforms Hydra-MDP, which follows VADv2's sampling-from-vocabulary paradigm, with a 5.1 PDMS improvement.
Even compared to the Hydra-MDP-$\mathcal{V}_{8192}$-W-EP, which is a variant of Hydra-MDP~\cite{hydramdp} by further training to fit the EP evaluation metric with additional supervision and using weighted confidence post-processing, \name{} still outperforms it by 3.5 EP and 1.6 overall PDMS, relying solely on a straightforward learning-from-human approach without any post-processing.
Compared to the Transfuser baseline, where we only differ in the planning module, \name{} delivers a notable 4.1 PDMS improvement, outperforming it across all sub-scores.

\subsection{Roadmap}
\begin{table*}[htbp!]
    \centering
    \renewcommand\tabcolsep{5.8pt}
    \input{tables/tab0}
    \vspace{-0.3cm}
    \caption{\textbf{Comparison on nuScenes dataset with open-loop metrics.} FPS is measured on a single NVIDIA 4090 GPU following the recipe of SparseDrive~\cite{sun2024sparsedrive}. Metric calculation follows ST-P3~\cite{hu2022st}.}
    \label{tab:main_nusc}
    \vspace{-0.6cm}
\end{table*}
In Tab.~\ref{tab:roadmap}, converting Transfuser into the generative Transfuser$_{\text{DP}}$ using vanilla diffusion policy improves the PDMS score by 0.6 and the mode diversity score $\mathcal{D}$ by 11\%.
However, it also significantly increases the overhead of the planning module, requiring 20$\times$ more denoising steps and 32$\times$ the step time, resulting in a total 650$\times$ increase in runtime overhead.
With the proposed truncated diffusion policy, Transfuser$_{\text{TD}}$ reduces the number of denoising steps from 20 to 2 while improving PDMS by 1.1 and mode diversity by 59\%.
By further incorporating the proposed diffusion decoder, the final model, \name{}, reaches  88.1 PDMS and 74\% mode diversity score $\mathcal{D}$.
Compared to the Transfuser$_{\text{DP}}$, \name{} shows improvements in 3.5 PDMS and  64\% mode diversity, and a 10$\times$ reduction in denoising steps, resulting in a 6$\times$ speedup in FPS. This enables real-time, high-quality, multi-mode planning.

\subsection{Ablation Study}

\boldparagraph{Effect of designs in diffusion decoder.}
Tab.~\ref{tab:design} shows the effectiveness of our design choices in the diffusion decoder.
ID-1 is the Transfuser$_{\text{TD}}$ in the Tab.~\ref{tab:roadmap}. By comapring ID-6 and ID-1, we can see that the proposed diffusion decoder reduce the 39\% parameters and significantly improves the planning quality by 2.4 PDMS.
ID-2 shows severe performance degeneration due to the lack of rich and hierarchical interaction with the environment.
By comparing ID-2 and ID-3, we show that spatial cross-attention is vital for accurate planning. ID-5 shows that the proposed cascade mechanism is effective and can further improve the performance.

\boldparagraph{Denoising step number.} Tab.~\ref{tab:step} shows that due to a reasonable start point, \name{} achieves good planning quality even with only 1 step, and further denoising steps offer quality improvement and inference flexibility for complex environments.

\boldparagraph{Cascade stages.}
Tab.~\ref{tab:stage} ablates the impact of cascade stage number. Increasing the stage number can improve the planning quality but saturate at the 4 stages and cost more parameters and inference time at each step.

\boldparagraph{Number of sampled noises $N_{\text{infer}}$.}
As stated in Sec.~\ref{sec:truncated}, \name{} can generate varied trajectories by simply sampling a variable number of noises from anchored Gaussian distribution.
Tab.~\ref{tab:sampled} shows that 10 sampled noises can already achieve a decent planning quality. By sampling more noises, \name{} can cover potential planning action space and lead to improved planning quality.

\subsection{Qualitative Comparison}
Since the PDMS planning metric calculates based on the top-1 scoring trajectory and our proposed $\mathcal{D}$ score evaluates mode diversity, these metrics alone cannot fully capture the quality of diverse trajectories.
To further validate the quality of multi-mode trajectories, we visualize the planning results of Transfuser, Transfuser$_{\text{DP}}$ and \name{} on challenging scenarios of NAVSIM \texttt{navtest} split in Fig.~\ref{fig:navsim_comp}.
The results indicate that the multi-mode trajectories generated by \name{} are not only diverse but also of high quality.
In Fig.~\ref{fig:navsim_comp_0}, the top-1 scoring trajectory generated by \name{} closely resembles the ground-truth trajectory, while the highlighted top-10 scoring trajectory surprisingly tries to perform high-quality lane changing.
In Fig.~\ref{fig:navsim_comp_1}, the highlighted top-10 scoring trajectory also performs a lane change, and a neighboring low-scoring trajectory further interacts with surrounding agents to effectively avoid collisions.

\subsection{Quantitative Comparison on nuScenes dataset}
The nuScenes dataset is previously popular benchmark for end-to-end planning. Since the major scenarios of nuScenes are simple and trivial situations, we only perform comparison in Tab.~\ref{tab:main_nusc}.
We implement \name{} on top of SparseDrive~\cite{sun2024sparsedrive} following its training and inference recipe using open-loop metrics proposed in ST-P3~\cite{hu2022st}. We stack 2 cascade diffusion decoder layers and apply the truncated diffusion policy with 18 clustered anchors.

As shown in Tab.~\ref{tab:main_nusc}, \name{} reduces the average L2 error of SparseDrive by 0.04m, achieving the lowest L2 error and average collision rate.
While \name{} is also efficient and runs 1.8$\times$ faster than VAD with 20.8\% lower L2 error and 63.6\% lower collision rate.

%% file: tables/tab0.tex
\begin{tabular}{l|cc|cccc|cccc|c}
    \toprule
    \multirow{2}{*}{Method} & \multirow{2}{*}{Input} & \multirow{2}{*}{Img. Backbone} &
    \multicolumn{4}{c|}{L2 (m) $\downarrow$} & 
    \multicolumn{4}{c|}{Collision Rate (\%) $\downarrow$} & \\
    && & 1s & 2s & 3s & \cellcolor{gray!30}Avg. & 1s & 2s & 3s & \cellcolor{gray!30}Avg. & \multirow{-2}*{FPS $\uparrow$}  \\
    \midrule
    ST-P3~\cite{hu2022st} & Camera & EffNet-b4~\cite{tan2019efficientnet} & 1.33 & 2.11 & 2.90 & \cellcolor{gray!30}2.11 & 0.23 & 0.62 & 1.27 & \cellcolor{gray!30}0.71 &  1.6 \\
    UniAD~\cite{hu2023planning} & Camera & ResNet-101~\cite{he2016deep}  & 0.45&0.70&1.04 & \cellcolor{gray!30}0.73 & 0.62&0.58&0.63 & \cellcolor{gray!30}0.61 & 1.8 \\
    OccNet~\cite{tong2023scene} & Camera & ResNet-50~\cite{he2016deep}  & 1.29 & 2.13 & 2.99 & \cellcolor{gray!30}2.14 & 0.21 & 0.59 & 1.37 & \cellcolor{gray!30}0.72  & 2.6 \\
    VAD~\cite{jiang2023vad}  & Camera & ResNet-50~\cite{he2016deep}  & 0.41&0.70&1.05 & \cellcolor{gray!30}0.72 & 0.07&0.17&0.41 & \cellcolor{gray!30}0.22 & 4.5 \\
    SparseDrive~\cite{sun2024sparsedrive}&Camera & ResNet-50~\cite{he2016deep} & 0.29 & 0.58 & 0.96 & \cellcolor{gray!30}0.61 & \textbf{0.01} & \textbf{0.05} & \underline{0.18} & \cellcolor{gray!30}\textbf{0.08} & \textbf{9.0}\\
    \midrule
    \name{} (Ours) & Camera & ResNet-50~\cite{he2016deep} & \textbf{0.27} & \textbf{0.54} & \textbf{0.90} & \cellcolor{gray!30}\textbf{0.57} & \underline{0.03} & \textbf{0.05} & \textbf{0.16} & \cellcolor{gray!30}\textbf{0.08} & \underline{8.2} \\
    \bottomrule
\end{tabular}%

%% file: sec/5_conclusion.tex
\section{Conclusion}
In this work, we propose a novel generative driving decision-making model, \name{}, for end-to-end autonomous driving by incorporating the proposed truncated diffusion policy and efficient cascade diffusion decoder. \name{} can denoise a variable number of samples from an anchored Gaussian distribution to generate diverse planning trajectories at real-time speeds. Comprehensive experiments and qualitative comparisons validate the superiority of \name{} in planning quality, running efficiency, and mode diversity.

%% file: sec/X_suppl.tex
\clearpage
\setcounter{page}{1}
\maketitlesupplementary
\appendix

\section{Further Implementation Detail}
We provide additional implementation details for our method on the NAVSIM~\cite{dauner2024navsim} and nuScenes~\cite{caesar2020nuscenes} datasets.

\paragraph{NAVSIM Dataset.}
We initialize the ResNet-34~\cite{he2016deep} backbone with ImageNet pre-trained weights, and the LiDAR range is 32m to the front, back, left, and right following Transfuser baseline~\cite{dauner2024navsim}.
We also perform auxiliary perception tasks following Transfuser baseline~\cite{dauner2024navsim}, which include 3D object detection, 2D BEV semantic segmentation. The object queries and BEV features are taken as input of the proposed diffusion decoder.

\paragraph{nuScenes Dataset.}
We follow the SparseDrive baseline~\cite{sun2024sparsedrive} to perform two-stage training. The model is directly initialized with the stage-1 pre-trained weight, which is trained solely on perception tasks (3D object detection/tracking, vectorized HD map construction, and motion prediction) and provided by the official open-source implementation. We train the stage-2 model on the nuScenes dataset for 10 epochs, replacing the planning module of SparseDrive with our proposed diffusion decoder and truncated diffusion mechanism. Object queries, map queries, and PV features are taken as inputs to the diffusion decoder.

\section{Further Ablation Study}
\begin{table}[h]
    \vspace{-4mm}
    \setlength{\tabcolsep}{4pt}
    \centering
    \resizebox{0.98\linewidth}{!}{
    \begin{tabular}{ll|ccccc|c}
        \toprule
        {Train} & Infer &NC$\uparrow$&DAC$\uparrow$&TTC$\uparrow$&Comf.$\uparrow$&EP$\uparrow$&PDMS$\uparrow$ \\
        \midrule
        \multirow{2}{*}{Anchored Dist.} &\cellcolor{blue!20}Anchored. Dist. &\cellcolor{blue!20}98.2&\cellcolor{blue!20}96.2&\cellcolor{blue!20}94.7&\cellcolor{blue!20}100&\cellcolor{blue!20}82.2&\cellcolor{blue!20}88.1\\
         & Extra. Traj.& 96.3 & 91.7 & 90.4 & 100 & 76.8 & 81.3  \\
        \midrule
        Extra. Traj. & Extra. Traj.& 97.3 & 94.0 & 92.6&100&79.6&84.7\\
        \bottomrule
    \end{tabular}}
    \vspace{-3.5mm}
    \caption{\textbf{Comparison on driving priors.} ``Anchored Dist.'' denotes anchored Gaussian distribution. ``Extra. Traj.'' denotes extrapolated trajectory based on current status. Row-1 marked in blue denotes the DiffusioDrive baseline of the main paper.}
    \label{tab:anchor_vs_traj}
    \vspace{-5.5mm}
\end{table}
\paragraph{Comparison on driving priors.}
In Tab.~\ref{tab:anchor_vs_traj}, we validate the superiority of prior anchors over the prior extrapolated trajectory based on the current status.
Row-1 is DiffusionDrive baseline. Row-2 uses the DiffusionDrive baseline model to infer from an extrapolated trajectory instead of sampled $N_{\text{infer}}$ trajectories.  Row-3 represents DiffusionDrive trained with a single anchor, \ie, the extrapolated trajectory, and infers by sampling around it. The results demonstrate the superiority of the proposed anchored Gaussian distribution over extrapolated prior, which fails to cover the potential action space and can not effectively handle challenging scenarios (\eg, obstacle avoidance and turning) in real-world application (consistent with comparisons to ego-status-based planners in Tab.~1 of NAVSIM paper~\cite{dauner2024navsim}).

\begin{table}[h]
    \vspace{-4mm}
    \setlength{\tabcolsep}{4pt}
    \centering
    \resizebox{1.0\linewidth}{!}{
    \begin{tabular}{lc|cccccc}
        \toprule
        Method & Anchor Source & \multicolumn{2}{c}{DS$\uparrow$} & \multicolumn{2}{c}{RC$\uparrow$}& \multicolumn{2}{c}{IS$\uparrow$}\\
        \midrule
        Transfuser$^{\dagger}$ & - &\multicolumn{2}{c}{47.30$_{\pm5.72}$} & \multicolumn{2}{c}{93.38$_{\pm1.20}$} & \multicolumn{2}{c}{0.50$_{\pm0.06}$} \\
        DiffusionDrive & NAVSIM & \multicolumn{2}{c}{\textbf{64.27$_{\pm2.43}$}} & \multicolumn{2}{c}{\textbf{94.16$_{\pm1.46}$}} & \multicolumn{2}{c}{\textbf{0.69$_{\pm0.02}$}} \\
        \bottomrule
    \end{tabular}}
    \vspace{-3.5mm}
    \caption{\textbf{Generalization of anchor source.} We test DiffusionDrive on Carla Longest6 benchmark with clustered anchors from NAVSIM dataset. $\dagger$ denotes that the result is taken from Transfuser paper~\cite{transfuser}.}
    \label{tab:anchor-transfer}
    \vspace{-5.5mm}
\end{table}

\paragraph{Generalization of anchor source.}
To further investigate the generalization of anchor source, we train DiffusionDrive on CARLA~\cite{dosovitskiy2017carla} with NAVSIM-clustered anchored Gaussian distribution (Row-2 in Tab.~\ref{tab:anchor-transfer}).
Since the CARLA dataset is totally different from NAVSIM, the superior results validate the generalization capability of our anchored Gaussian distribution, which is designed to cover potential multi-mode driving action space instead of train/val information leakage.

\section{Further Qualitative Comparison}
\begin{figure*}[ht!]
    \centering
    \begin{subfigure}[b]{0.82\linewidth}
        \includegraphics[width=\linewidth]{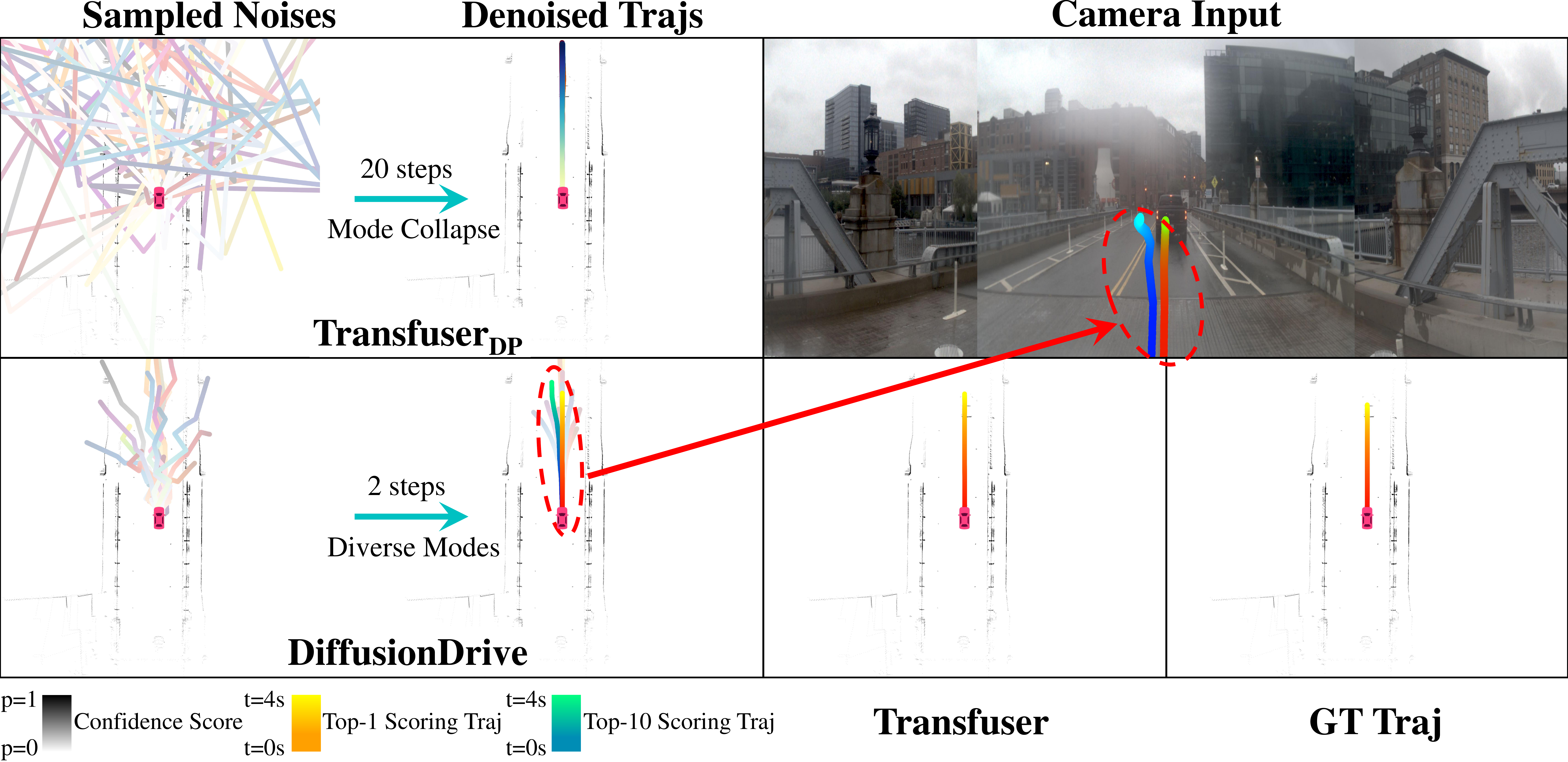}
        \vspace{-0.6cm}
        \caption{.}
        \label{fig:straight_comp_0}
    \end{subfigure}
    \begin{subfigure}[b]{0.82\linewidth}
        \includegraphics[width=\linewidth]{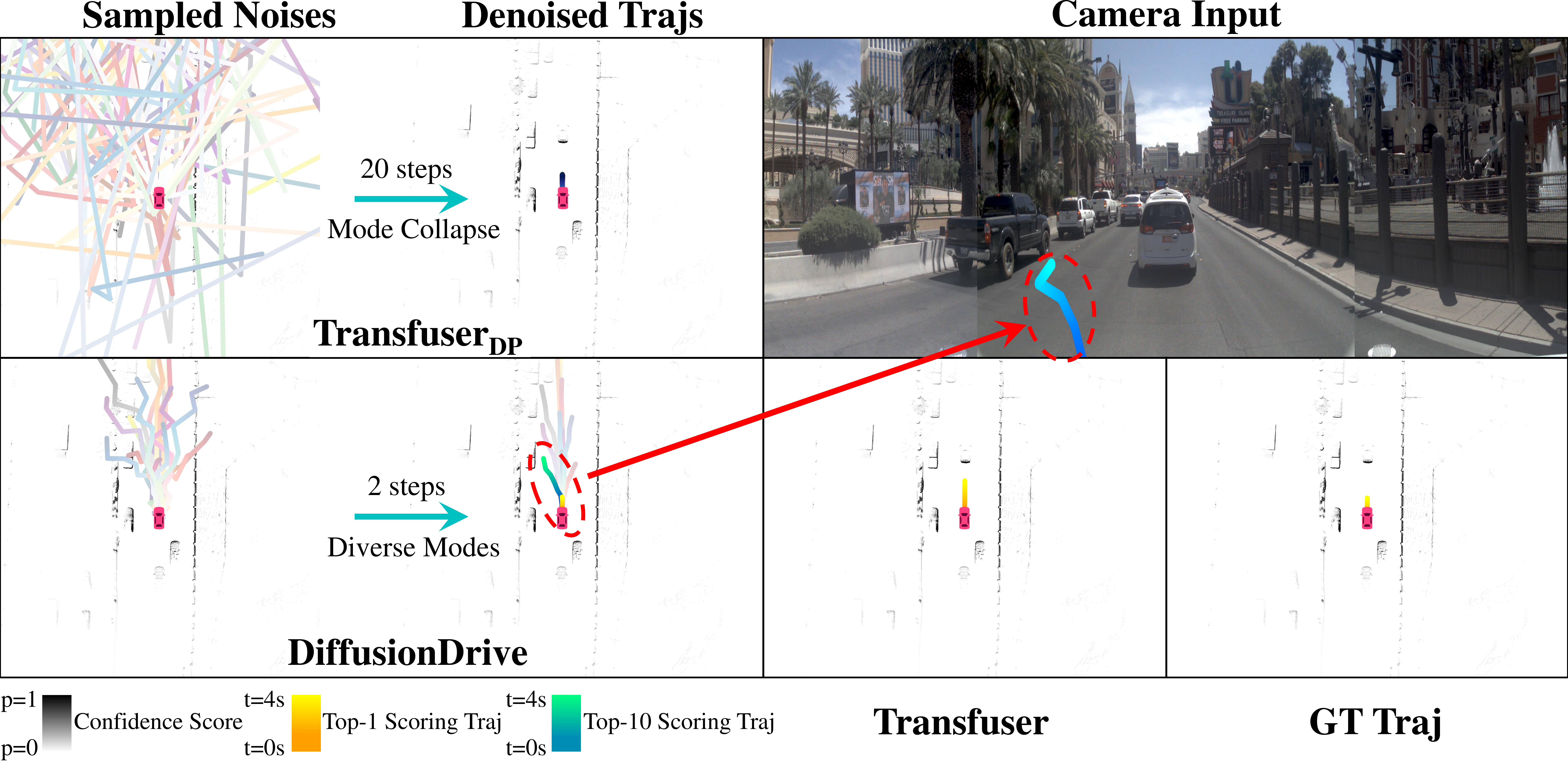}
        \vspace{-0.6cm}
        \caption{.}
        \label{fig:straight_comp_1}
    \end{subfigure}
    \begin{subfigure}[b]{0.82\linewidth}
        \includegraphics[width=\linewidth]{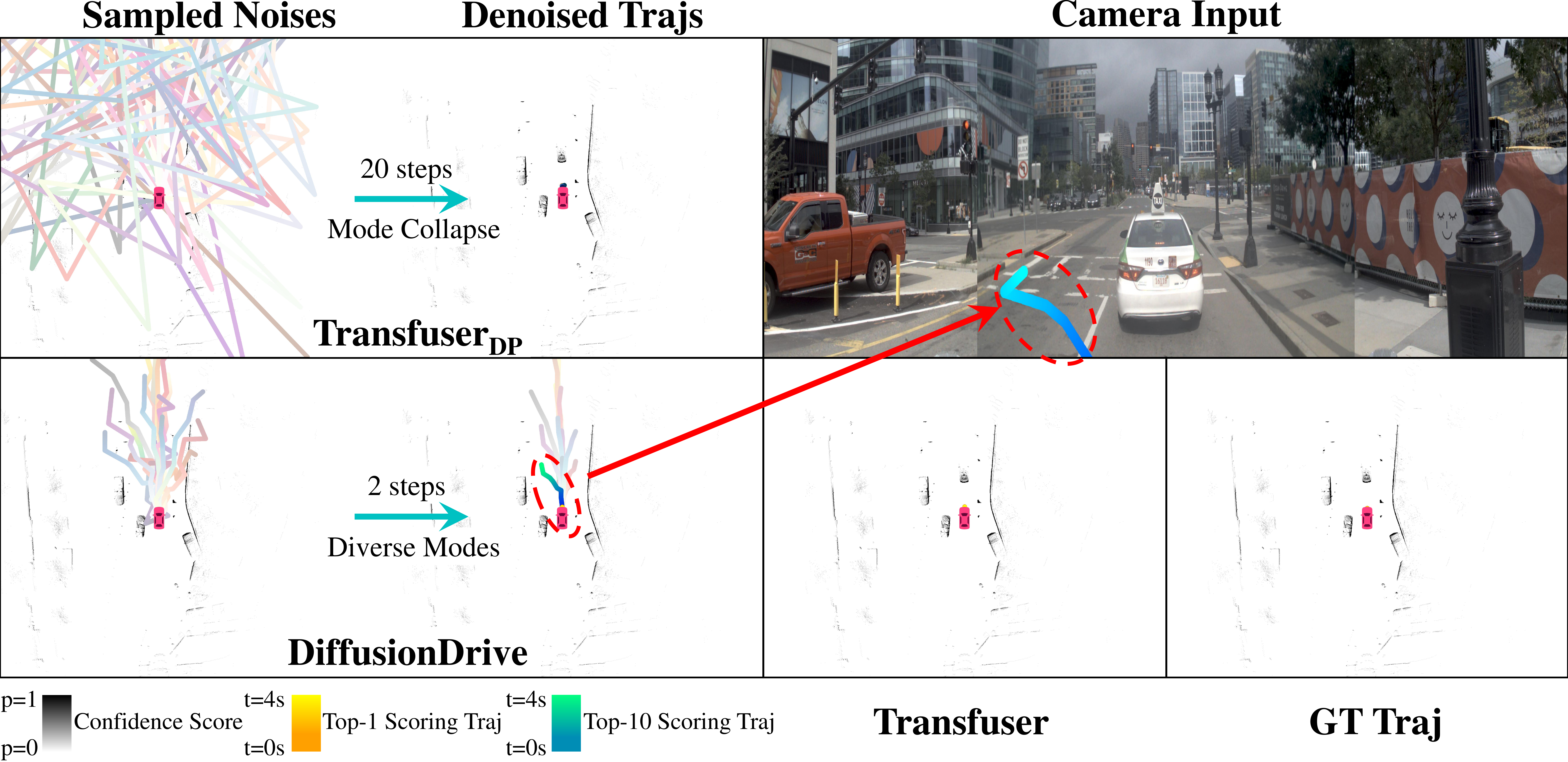}
        \vspace{-0.6cm}
        \caption{.}
        \label{fig:straight_comp_2}
        \vspace{-0.4cm}
    \end{subfigure}
    \caption{\textbf{Qualitative comparison of Transfuser, Transfuser$_{\text{DP}}$ and \name{} on going straight scenarios of NAVSIM \texttt{navtest} split.}}
    \label{fig:supp_comp_straight}
\end{figure*}

In this section, we provide additional qualitative comparisons on challenging scenarios from the planning-oriented NAVSIM dataset \texttt{navtest} split~\cite{dauner2024navsim}.
\paragraph{Going straight.}
Fig.~\ref{fig:straight_comp_0} and Fig.~\ref{fig:straight_comp_1} show that the top-1 scoring trajectories of \name{} are similar to the ground truth trajectories, while the highlighted top-10 scoring trajectories can perform robust lane changes. Notably, Fig.~\ref{fig:straight_comp_2} demonstrates that the diverse and highlighted top-10 trajectories can further recognize the traffic light, enabling reasonable lane changes and stopping at the stop line.

\begin{figure*}[ht!]
    \centering
    \begin{subfigure}[b]{0.82\linewidth}
        \includegraphics[width=\linewidth]{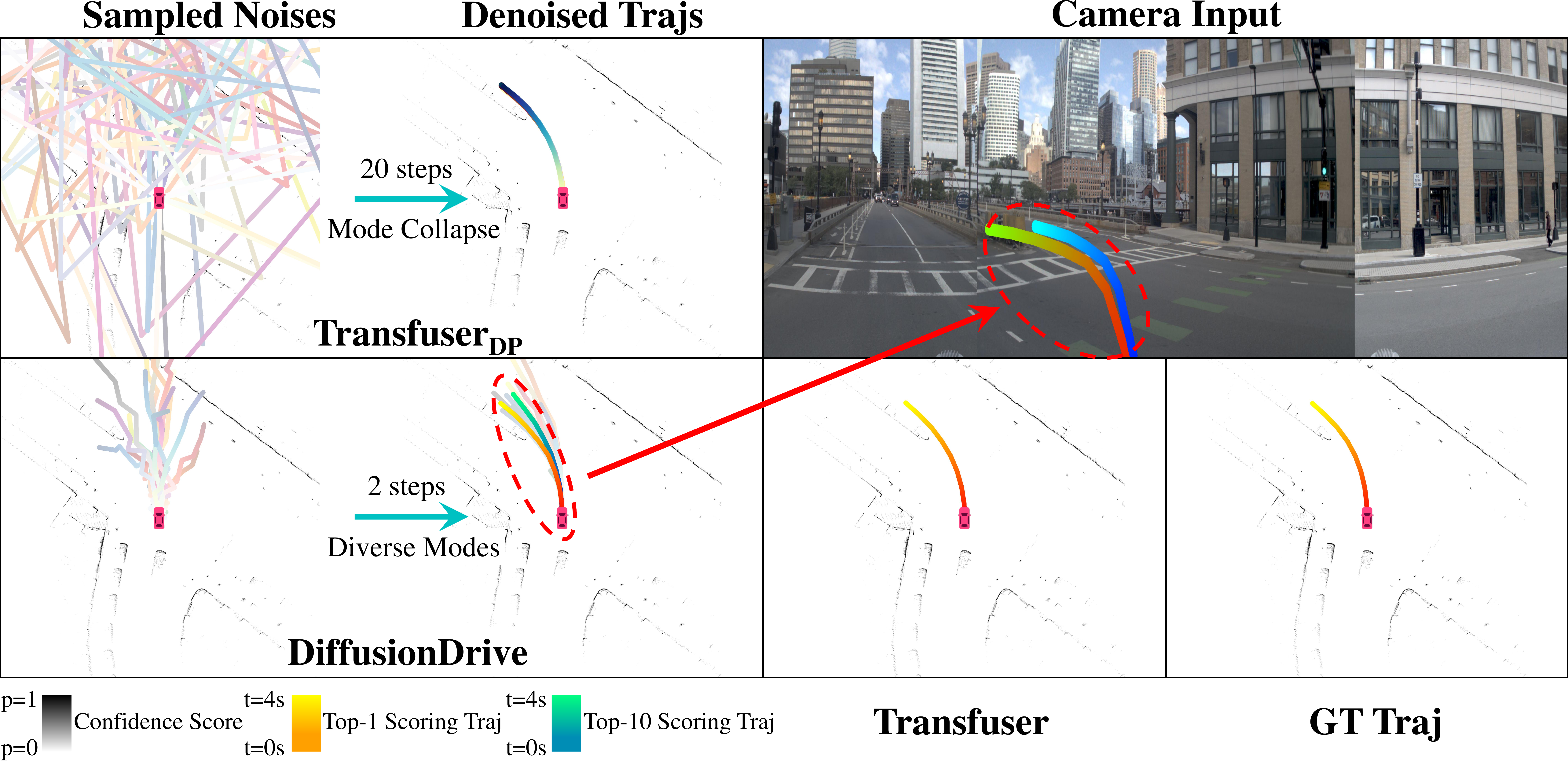}
        \vspace{-0.6cm}
        \caption{.}
        \label{fig:left_comp_0}
    \end{subfigure}
    \begin{subfigure}[b]{0.82\linewidth}
        \includegraphics[width=\linewidth]{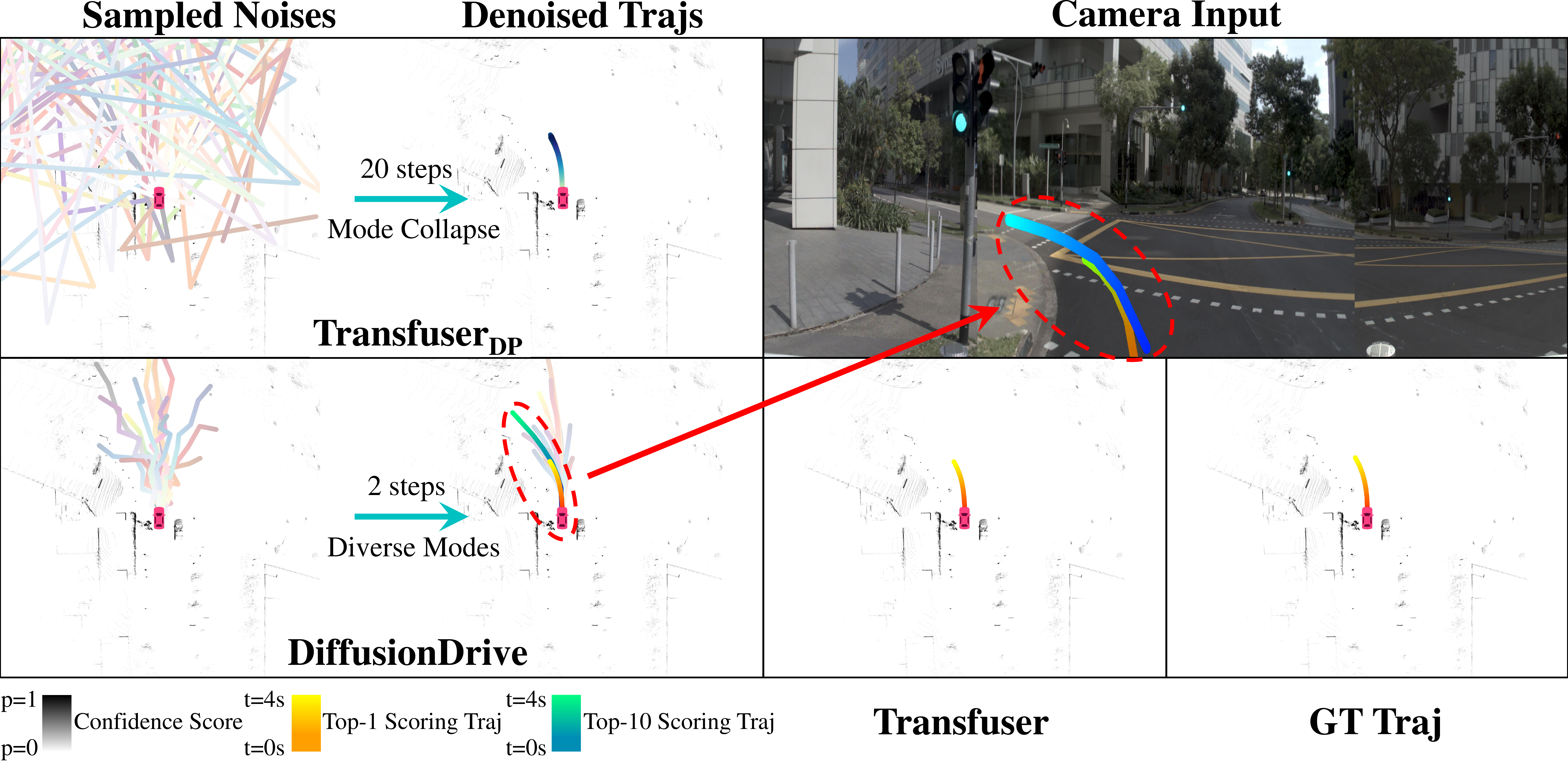}
        \vspace{-0.6cm}
        \caption{.}
        \label{fig:left_comp_1}
    \end{subfigure}
    \begin{subfigure}[b]{0.82\linewidth}
        \includegraphics[width=\linewidth]{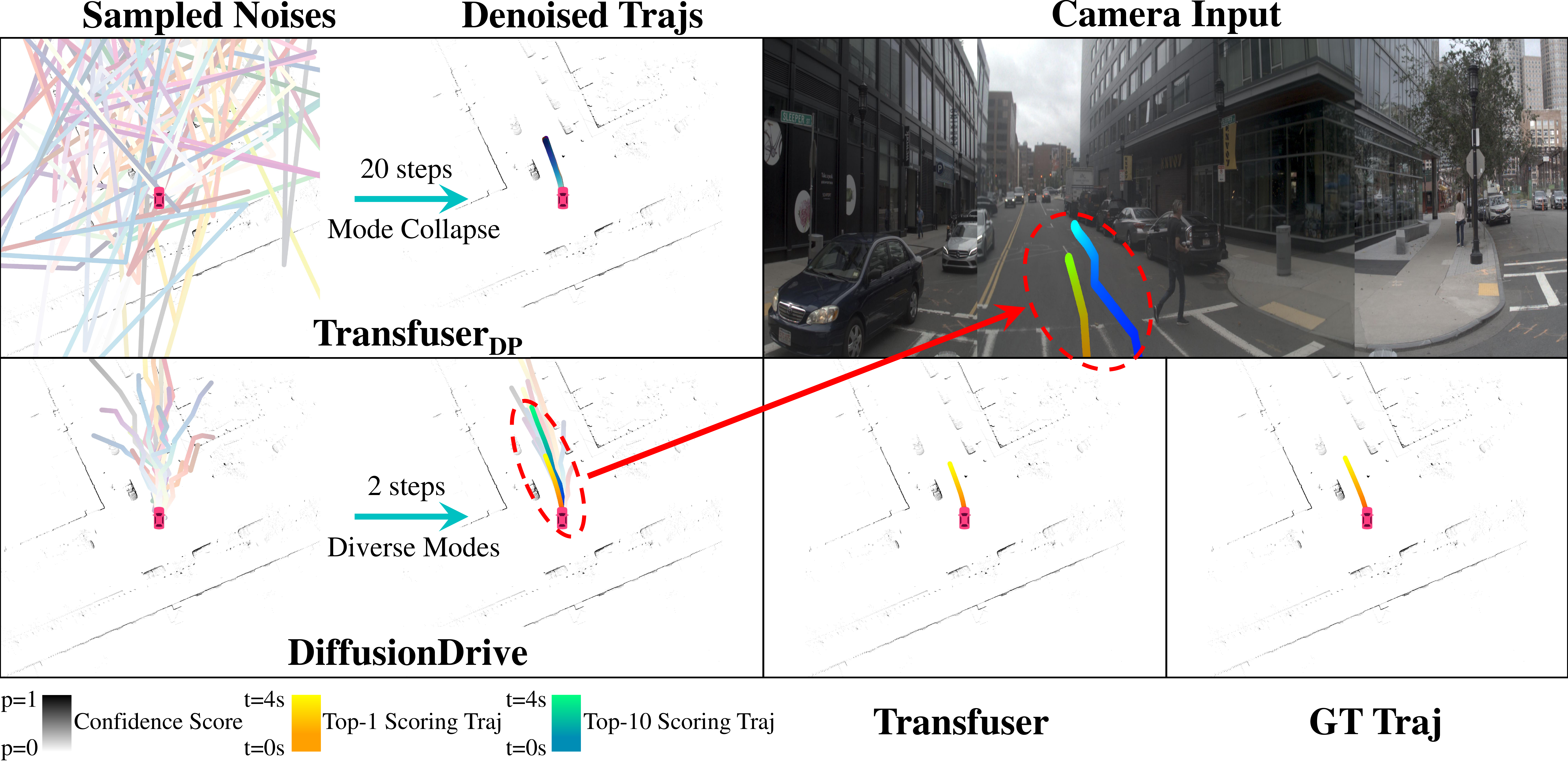}
        \vspace{-0.6cm}
        \caption{.}
        \label{fig:left_comp_2}
        \vspace{-0.4cm}
    \end{subfigure}
    \caption{\textbf{Qualitative comparison of Transfuser, Transfuser$_{\text{DP}}$ and \name{} on turning left scenarios of NAVSIM \texttt{navtest} split.}}
    \label{fig:supp_comp_left}
\end{figure*}

\paragraph{Turning left.}
Fig.~\ref{fig:supp_comp_left} shows that the denoised diverse trajectories are dynamically adjusted based on the traffic conditions. The highlighted top-10 scoring trajectories are robust and reasonable, effectively performing lane changes.

\begin{figure*}[ht!]
    \centering
    \begin{subfigure}[b]{0.82\linewidth}
        \includegraphics[width=\linewidth]{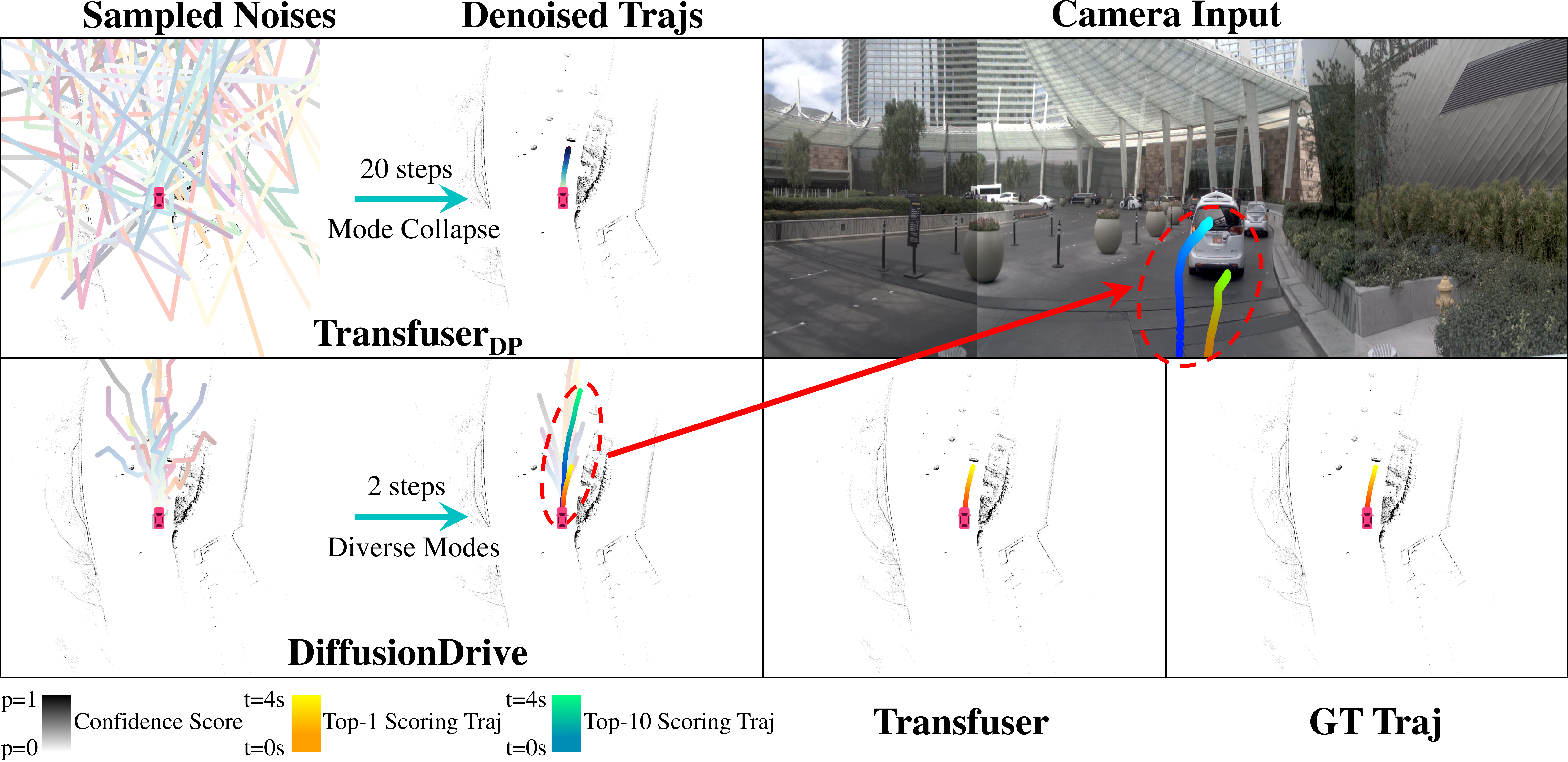}
        \vspace{-0.6cm}
        \caption{.}
        \label{fig:right_comp_0}
    \end{subfigure}
    \begin{subfigure}[b]{0.82\linewidth}
        \includegraphics[width=\linewidth]{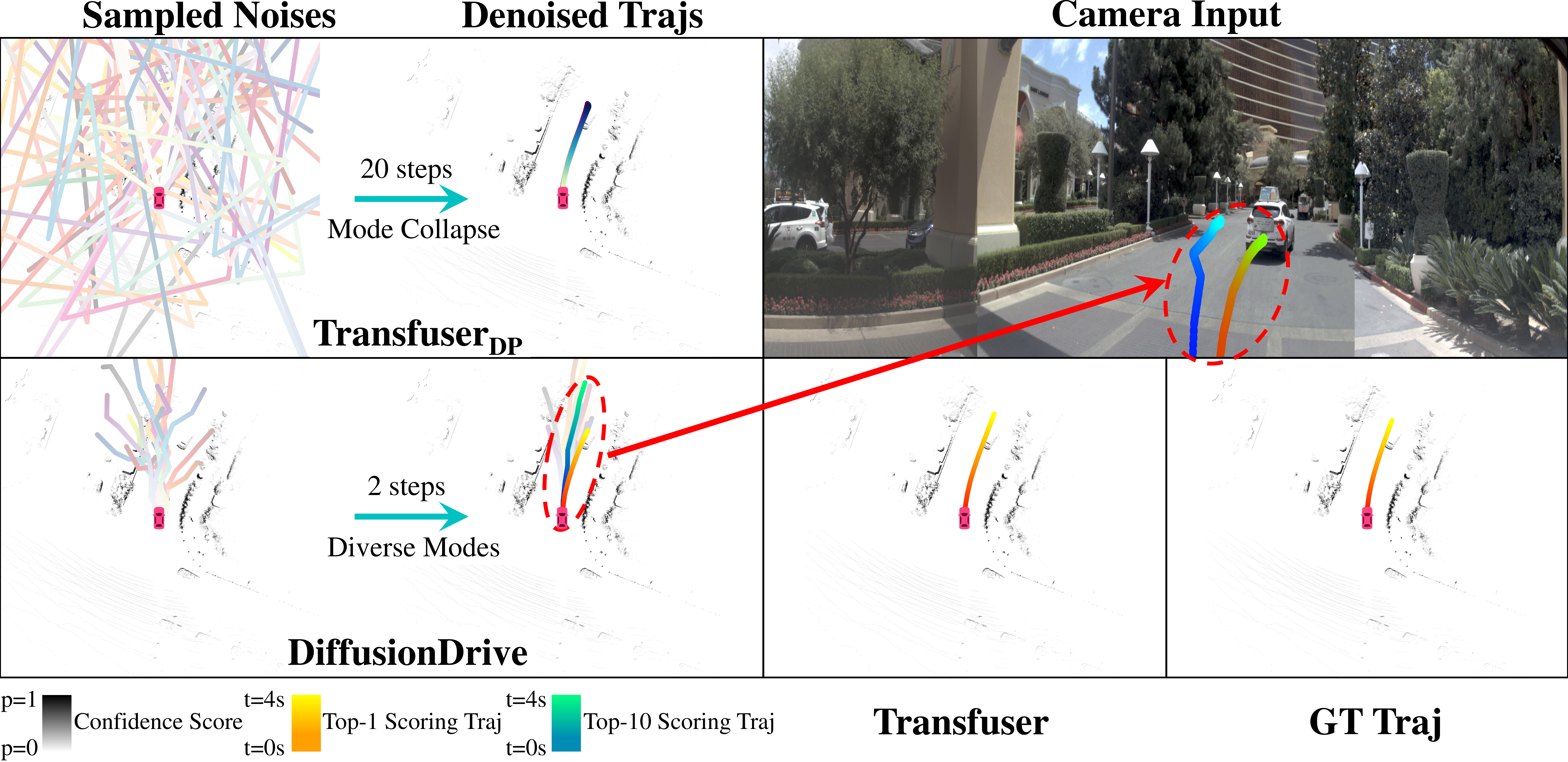}
        \vspace{-0.6cm}
        \caption{.}
        \label{fig:right_comp_1}
    \end{subfigure}
    \begin{subfigure}[b]{0.82\linewidth}
        \includegraphics[width=\linewidth]{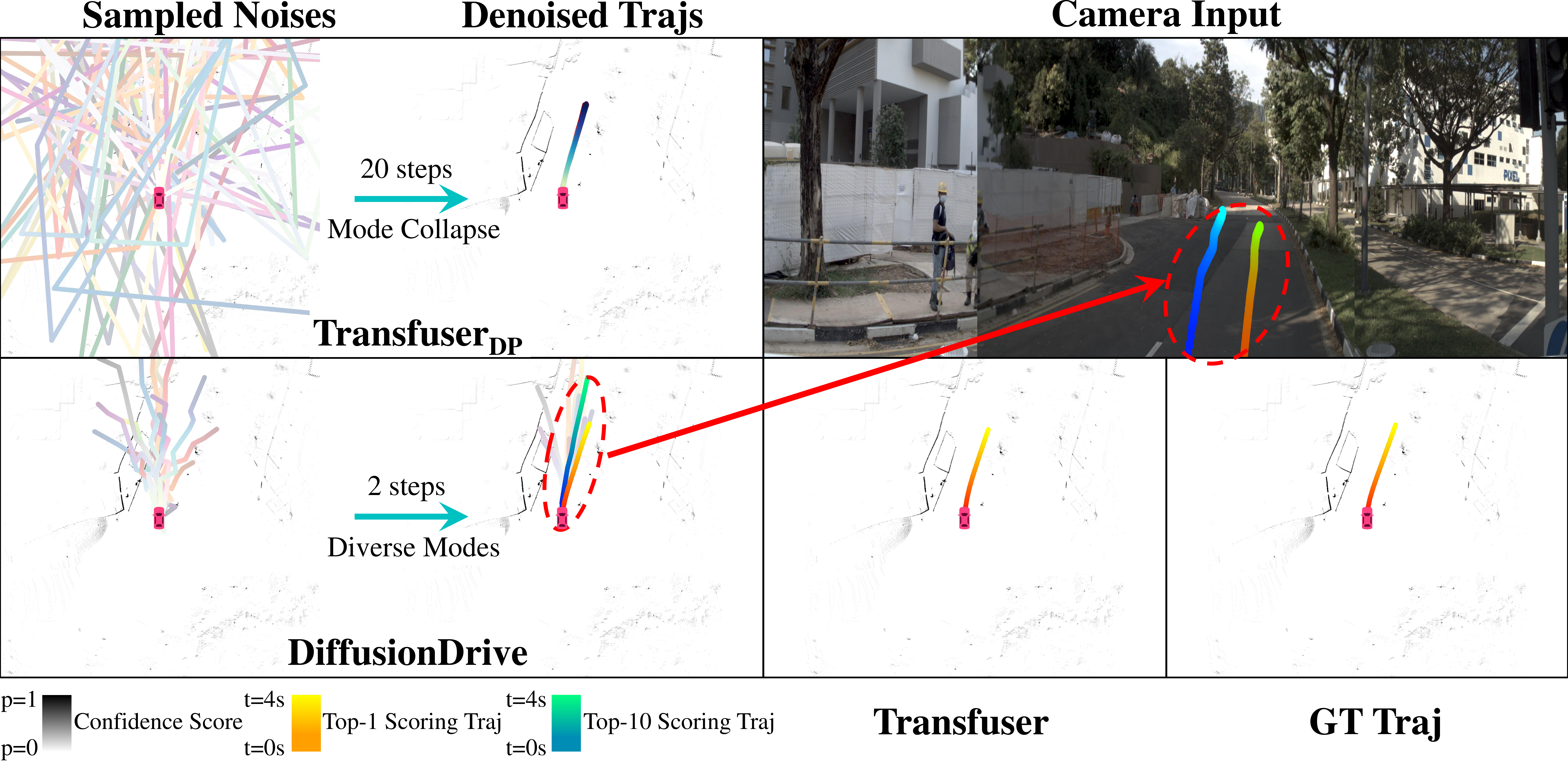}
        \vspace{-0.6cm}
        \caption{.}
        \label{fig:right_comp_2}
        \vspace{-0.4cm}
    \end{subfigure}
    \caption{\textbf{Qualitative comparison of Transfuser, Transfuser$_{\text{DP}}$ and \name{} on turning right scenarios of NAVSIM \texttt{navtest} split.}}
    \label{fig:supp_comp_right}
\end{figure*}

\paragraph{Turning right.}
Fig.~\ref{fig:right_comp_0} and Fig.~\ref{fig:right_comp_1} show that the top-1 scoring trajectories of \name{} are going to perform car-following like the ground truth trajectories, while the highlighted top-10 scoring trajectories tend to overtake the leading vehicle.
These results validate that \name{} can robustly generate diverse and plausible driving actions.

